%% file: paper.tex
\documentclass{article} 
\usepackage{iclr2025_conference,times}

\newcommand{\ourmaintitle}{The Uncanny Valley: Exploring Adversarial Robustness from a Flatness Perspective}
\iclrfinalcopy

\makeatletter
\newif\if@restonecol
\makeatother

\usepackage{latexsym}
\usepackage[ruled, vlined, nofillcomment, linesnumbered]{algorithm2e}
\usepackage[notheorems]{eda} 
\usepackage{bbm}
\usepackage{caption}
\usepackage{subcaption}
\usepackage[pdftitle={ourtitle}, pdfauthor={Double Blind Submission}, hidelinks, hyperfootnotes = false]{hyperref}
\usepackage{wrapfig}
\graphicspath{ {./figs/} }
\usepackage{prettyplots} 

\input{defines}
\input{commands}

\input{math_commands.tex}

\usepackage{hyperref}
\usepackage{url}

\title{\ourmaintitle}

\author{
  Nils Philipp Walter\\
  CISPA Helmholtz Center \\
  for Information Security\\
  \texttt{nils.walter@cispa.de}\\
  \And
  Linara Adilova\\
  Ruhr University Bochum\\
  \texttt{linara.adilova@rub.de}\\
  \And
  Jilles Vreeken\\
  CISPA Helmholtz Center\\
  for Information Security\\
  \texttt{vreeken@cispa.de}\\
  \And
  Michael Kamp\\
  Institute for AI in medicine (IKIM), University Hospital Essen\\
  and Ruhr University Bochum and Monash University\\
  \texttt{michael.kamp@uk-essen.de} \\
}

\begin{document}

\maketitle

\begin{abstract}
   \input{abstract}

\end{abstract}
\input{introduction}

\input{related}

\input{preliminaries}

\input{empiricalRelation}

\input{experiments}
\input{theory}

\input{discussion}

\bibliographystyle{iclr2025_conference}
\bibliography{bib/abbrev,bib/bib-jilles,bib/bib-paper, bib/related}

\appendix

\include{appendix}

\end{document}

%% file: commands.tex
\newcommand{\pred}{ \hat{y} \xspace}
\newcommand{\partl}[1]{\frac{\partial}{\partial w_{#1}} \xspace}
\DeclareMathOperator{\pro}{\Pi}

\newcommand{\resnet}{\textsc{ResNet-18}\xspace}
\newcommand{\wrn}{\textsc{WideResNet-28-4}\xspace}
\newcommand{\dense}{\textsc{DenseNet121}\xspace}
\newcommand{\vgg}{\textsc{VGG11}\xspace}

\newcommand{\llama}{\textsc{LLama2}\xspace}
\newcommand{\vicuna}{\textsc{Vicuna}\xspace}
\newcommand{\guanaco}{\textsc{Guanaco}\xspace}

%% file: math_commands.tex
\usepackage{amsmath,amsfonts,bm}

\def\eqref#1{equation~\ref{#1}}

\def\1{\bm{1}}

\DeclareMathAlphabet{\mathsfit}{\encodingdefault}{\sfdefault}{m}{sl}
\SetMathAlphabet{\mathsfit}{bold}{\encodingdefault}{\sfdefault}{bx}{n}

\newcommand{\R}{\mathbb{R}}

\DeclareMathOperator*{\argmax}{arg\,max}

%% file: abstract.tex
Flatness of the loss surface not only correlates positively with generalization, but is also related to adversarial
robustness, since perturbations of inputs relate non-linearly to perturbations of weights.
In this paper, we empirically analyze the
relation between adversarial examples and relative flatness with respect to the parameters of one layer. We observe a peculiar
property of adversarial examples in the context of relative flatness: during an iterative first-order white-box attack, the flatness of the loss surface measured around
the adversarial example \emph{first} becomes sharper until the label is flipped, but if we keep the attack running, it runs into a flat \textit{uncanny valley} where the label remains 
flipped. In extensive experiments, we observe this phenomenon across various model architectures and datasets, even for adversarially trained models. Our results also extend to large language models (LLMs),
but due to the discrete nature of the input space and comparatively weak attacks,
adversarial examples rarely reach truly flat regions. Most importantly, this phenomenon shows that flatness alone cannot explain adversarial robustness unless we can also guarantee the behavior of the function around the examples. We therefore theoretically connect relative flatness
to adversarial robustness by bounding the third derivative of the loss surface, underlining the need for flatness in combination with a low global Lipschitz constant for a robust model.

%% file: introduction.tex
\section{Introduction}
\label{sec:intro}

Despite the remarkable performance of modern deep learning models, their vulnerability to adversarial attacks, i.e., small crafted perturbations in the input
that fool a model into changing its prediction to an incorrect label~\citep{szegedy2014intriguing, carlini2017adversarial}, undermine the trust in the
reliability of these models. Although there has been significant progress in developing methods to enhance adversarial robustness, the underlying mechanisms
that dictate the susceptibility of a model to such perturbations are still not fully understood.
One promising approach is the study of flatness of the loss surface. A lot of previous work demonstrated a positive correlation between flatness of the\input{latex_figs/intro-plot}
loss surface and the generalization ability of a model~\citep{hochreiter1994simplifying,keskar2016large, jiang2019fantastic}.
Flat minima in the loss landscape are thought to be indicative of better generalizing models, as they suggest that small changes in the parameters space do not significantly affect the loss of a model. This concept has led to the hypothesis that flatness is also related to adversarial robustness.
While flatness measured with respect to inputs obviously relates to adversarial robustness~\citep{moosavi2019robustness}, it is not clear how flatness with respect to parameters connects to it~\citep{wu2020adversarial, kanai2023relationship}.

In this paper, we explore the relationship between flatness of the loss surface with respect to parameters and adversarial examples. We empirically
investigate this connection by analyzing the behavior of the loss surface during iterative first-order white-box attacks. Our findings reveal an {intriguing} 
phenomenon, of which we give an example in Fig.~\ref{fig:uncanny}. The loss surface initially becomes sharper as the attack progresses and the prediction is flipped. However, if the attack is continued, the adversarial example often moves into a flat region of the loss surface, which we term \textit{uncanny valley}. This region
is uncanny in the sense that the surface around adversarial examples is very flat while they are still changing the prediction of the model.

This does not only indicate that the correlation between flatness and strong adversarial examples is not intuitive, but also that a vicinity of such an adversarial sample will be filled with similar adversarial examples.
We observe this uncanny valley phenomenon across various model architectures, including large language models (LLMs), datasets,
and also in adversarially trained models. Interestingly,  for more robust models, much stronger attacks are necessary to find these uncanny valleys. For LLMs, we find that their discrete nature and the availability of only relatively weak attacks often prevent adversarial examples from reaching truly flat
regions. 

This observation suggests that flatness alone cannot fully explain adversarial robustness; instead it is crucial to locally 
control the smoothness of the loss Hessian around the adversarial examples. Based on this insight, 
we derive a connection between relative flatness~\citep{petzka2021relative} and adversarial robustness by
bounding the third derivative of the loss surface. This bound provides a theoretical guarantee on adversarial robustness, linking the flatness of the loss landscape to
the robustness of a model to adversarial attacks.
Intuitively, the bound states that adversarial robustness increases as the network becomes flatter when the model function is simultaneously sufficiently smooth in the Lipschitz sense, i.e., \emph{small} Lipschitz constant.

This work highlights the complexity of adversarial robustness and emphasizes the need for a deeper understanding of the geometry of the loss landscape. By bridging the gap between flatness and adversarial robustness, we aim to pave the way for more robust and reliable deep learning models. Additionally our work elucidates the counter-intuitive behaviour of flatness measures based on the trace of the Hessian, when evaluated on adversarial examples.

In summary, our three main contributions are:
\begin{enumerate}
    \item We introduce and empirically demonstrate the phenomenon of the \textit{uncanny valley}, which is a plateau in the loss surface that we find via adversarial attacks.
    \item We perform this analysis on various model architectures and datasets, including convolutional neural networks (CNNs) and large language models (LLMs).
    \item We provide a theoretical framework that connects flatness to adversarial robustness through the third derivative of the loss surface.
\end{enumerate}

%% file: latex_figs/intro-plot.tex
\begin{wrapfigure}{R}{0.38\textwidth}
 \vspace{-0.5cm}
    \begin{subfigure}[t]{\linewidth}
            \begin{tikzpicture}
                    \usetikzlibrary{calc}
                    \begin{axis}[ 
                        pretty line,
                        cycle list name = prcl-line,
                        legend style={yshift=1.0cm,font=\scriptsize,xshift=1cm},
                        width=\linewidth,
                        height=3.25cm,
                        ylabel={Relative sharpness},
                        xlabel={Attack iteration},
                        pretty labelshift,
                        line width=0.75,
                        legend columns = 9,
                        xtick = {1, 3, 5, 7, 9, 11},
                        tick label style={/pgf/number format/fixed},
                    ]
                    \foreach \x in {wrn}{
                        \addplot table[x={0}, y={trace-\x-cifar10}, col sep=comma]{expres/normalized_rfm-for-clean.csv};
                    }    
                    \end{axis}
                \end{tikzpicture}
    \end{subfigure}
    \vspace{0.3em}
    \begin{subfigure}[t]{\linewidth}
            \begin{tikzpicture}
                    \usetikzlibrary{calc}
                    \begin{axis}[ 
                        pretty line,
                        cycle list name = prcl-line,
                        legend style={yshift=1.0cm,font=\scriptsize,xshift=6cm},
                        width=\linewidth,
                        height=3.25cm,
                        ylabel={Loss},
                        xlabel={Attack iteration},
                        pretty labelshift,
                        line width=0.75,
                        legend columns = 9,
                        xtick = {1, 3, 5, 7, 9, 11},
                        ytick = {0, 10, 20, 30, 40},
                        yticklabels = {0,10,20,30,\hphantom{0}40,50},
                        ymin=0,
                        ymax=40,
                        tick label style={/pgf/number format/fixed},
                    ]
                    \foreach \x in {wrn}{
                        \addplot+ table[x={0}, y={loss-\x-cifar10},col sep=comma]{expres/normalized_rfm-for-clean.csv};%
                    }    
                    \end{axis}
                \end{tikzpicture}
    \end{subfigure}%
    \caption{\textit{The Uncanny Valley}. During a multi-step adversarial attack, sharpness first increases; then decreases to almost zero (top), while the loss steadily increases (bottom).\label{fig:uncanny}}
    \vspace{-0.2cm}
\end{wrapfigure}
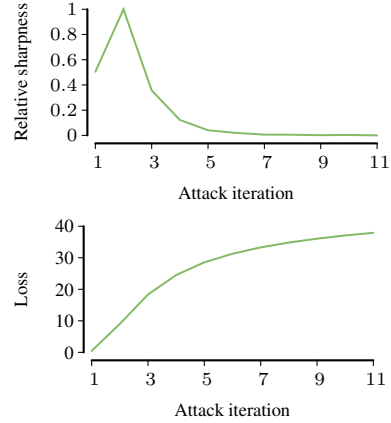%

%% file: related.tex
\section{Related Work}\label{sec:related}
\paragraph{Adversarial Examples}
\citet{szegedy2014intriguing} introduced the notion of adversarial inputs for deep learning models as a research subject, characterizing these as a curious property of neural networks.
High-level definition of an adversarial example is a perturbation of a benign input that is hard to detect for a human but which leads to the mistake in the model prediction. 
By now, they have become a major drawback of practical deep learning, since they not only pose a possible threat in applications, but they undermine the trust in machine learning in general: How can we trust the predictions of models that are so easily fooled?

In the meantime, many algorithms for generating adversarial examples have been proposed, starting with FGSM~\citep{goodfellow2014explaining}, followed by the defacto standard attacks PGD~\citep{madry2017towards} and C\&W~\citep{carlini2017towards}, and many more~\citep{papernot2016limitations, kurakin2017adversarial, narodytska_simple_2017, brown_adversarial_2018, alaifari2018adef, andriushchenko2020square, croce2020reliable, croce2021mind}.
With the development of generative deep learning, the approaches for generating adversarial samples with diffusion models~\citep{chen2023advdiffuser} appeared.
Large language models are not an exception and susceptible not only to standard input manipulation for prediction failure~\citep{li2020bert, zou2023universal} but also to so-called jailbreaks, which are forcing the model to output text that was supposed to be excluded from the possible results of the inference~\citep{wei2024jailbroken}.

\paragraph{Defenses against Adversarial Attacks}
The most common way to make a model more robust to adversarial examples is adversarial training, which incorporates adversarial inputs into the training procedure~\citep{szegedy2014intriguing, shafahi2019adversarial, kumari2019harnessing, perolat_playing_2018, shafahi2020universal, cai2018curriculum, tramer2018ensemble, wu2020adversarial, carmon2019unlabeled}.
It is usually observed that adversarial training reduces the performance of a model on clean data~\citep{tsipras2018robustness}.
There also exists contradictory research stating that adversarial training approaches lead to  a flatter loss surface with respect to the parameters~\citep{wu2020adversarial, stutz_relating_2021}. This, in turn, is believed to lead to better generalization~\citep{hochreiter1994simplifying}.

\citet{rahnama2020robust} propose an approach where each of the subnetworks corresponding to layers is robustified via insights from Lyapunov theory,
connecting robustness to spectral regularization.
Nevertheless, later \citet{liang2021large} show that large norms of layers do not always induce large global Lipschitz constant.
So regularization of norms is not an effective way to enhance robustness.
\citet{moosavi2019robustness} propose to penalize the Hessian with respect to input to improve the flatness in the input space and demonstrate that it improves adversarial robustness, analogous work was performed by \citet{xu2020adversarial}.

Interestingly, \citet{kanai2023relationship} analyses smoothness in the input space and smoothness in the parameter space and concludes that smoothness in the input space leads to a non-flat surface with respect to parameters, which leads to worse performance.
Flatness of the loss surface as a form of smoothness in parameter space has also been linked to generalization~\citep{liang2019fisher, tsuzuku2020normalized}.
\citet{foret2020sharpness} proposed Sharpness Aware Minimization (SAM), that is essentially similar to the work of \citet{wu2020adversarial}, but focuses on improving clean accuracy. The approach became very popular due to the ability to improve some of the state-of-the-art results in image classification.
\citet{dinh2017sharp} show, however, that flatness in terms of the loss Hessian of the entire network cannot predict generalization, due to reparameterizations.
\citet{petzka2021relative} showed that generalization instead can be linked to a relative flatness of a single layer of a network.

\paragraph{Lipschitz Continuity and Adversarial Robustness}
One of the drawbacks of adversarial training is the absence of guarantees on the robustness of the obtained model.
One of the popular ways to obtain guarantees is randomized smoothing~\citep{cohen2019certified}, which implicitly controls the global Lipschitz constant~\citep{salman2019provably}.
The main idea here is to produce a smoothed version of a classifier by making it predict the most probable label in a normal distribution surrounding an example. 
Directly using Lipschitz properties is not quite practical however for modern networks. Global Lipschitz constants~\citep{tsuzuku2018lipschitz} are usually too vague, while local Lipschitz constants~\citep{hein2017formal} are impractical to compute.
\citet{liang2021large} show that a large global Lipschitz does not make the local constants large, which means that even models with a large global Lipschitz constant can be adversarially robust.
Small global Lipschitz constant on the other hand helps to control local constants, but sacrifices clean accuracy.
\citet{yang2020closer} showed that improving local Lipschitz constants and enforcing better generalization can serve as a method to get good generalizing and robust models~, but it might not be easily achievable.

%% file: preliminaries.tex
\section{Preliminaries}
\label{sec:prelim}
We assume a distribution $\D$ over an input space $\X$ and a target space $\Y$ with corresponding probability density function $P(X,Y)=P(Y\mid X)P(X)
$, and models $f:\X\rightarrow\Y$ from a model class $\F$ and loss functions $\loss:\Y\times\Y\rightarrow\R_+$.
Intuitively, an adversarial example is a small and imperceptible perturbation $r^*$ of a sample $x$, such that the model 
incorrectly classifies the perturbed sample $\xi=x+r^*$. Formally, adversarial examples are defined as an optimization problem:
\begin{df}[\citet{szegedy2014intriguing, papernot2016limitations} and \citet{carlini2017towards}]
Let $f:\R^m\rightarrow\{1,\dots,k\}$ be a classifier, $x\in [0,1]^m$, and $l\in [k]$ with $l\neq f(x)$ a target class. Then for every
\[
r^* = \arg\min_{r\in\R^m}\|r\|_2\text{ s.t. }f(x+r)=l\text{ and }x+r\in [0,1]^m
\]
the perturbed sample $\xi=x+r^*$ is called an \textbf{adversarial example}. 
\label{def:advEx-classic}
\end{df}
The implicit assumptions here are that (i) the small size of the perturbation in $L_2$-distance of an example $x$ make the adversarial example $\xi$ {hard (e.g., for a human) to distinguish} from the original $x$ and (ii) the semantics are not altered in the sense that $y=f(x)$ is the true label and for the adversarial example $\xi$ the true label is still $y$. This means that the loss of an 
adversarial robust classifier $f^*$ on an adversarial example $\xi$ must not
increase significantly, i.e., $ l(f^*(\xi),y) - l(f^*(x),y) < \epsilon$. To ensure these assumptions are made explicitly and to improve clarity,
adversarial examples can be defined more generally as follows:
\begin{df}
Let $\D$ be a distribution over an input space $\X$ and a label space $\Y$ with corresponding probability density function $P(X,Y)=P(Y\mid X)P(X)$.
Let $\loss:\Y\times\Y\rightarrow\R_+$ be a loss function, $f\in\F$ a model, and $(x,y)\in\X\times\Y$ be an example drawn according to $\D$. Given a distance function $d:\X\times\X\rightarrow\R_+$ over $\X$ and two thresholds $\epsilon,\delta\geq 0$, we call $\xi\in\X$ an \defemph{adversarial example} for $x$ if $d(x,\xi)\leq\delta$ and
\[
\expec{y_\xi\sim P(Y\mid X=\xi)}{\loss(f(\xi),y_\xi)} - \loss(f(x),y) > \epsilon \; .
\]
\label{def:advEx}
\end{df}
By making the explicit assumptions that (i) the distance is measured via the $L_2$-norm, i.e., $d(x,\xi)=\|x-\xi\|_2$, (ii) $f(x)$ is correct, i.e., $f(x) = y$, and (iii) the true label for $\xi$ under $\D$ is $y$, i.e., $\expec{y_\xi\sim P(Y\mid X)}{\loss(f(\xi),y_\xi)}=\loss(f(\xi),y)$, this definition is equivalent to the original Def.~\ref{def:advEx-classic}. We prove that Def.~\ref{def:advEx} is a generalization of the classical Def.~\ref{def:advEx-classic} in Appendix~\ref{appdx:proof:defsAreEquivalent}.

Def.~\ref{def:advEx} naturally reminds of the ($\epsilon$, $\delta$)-criterion for continuity. Intuitively, it exemplifies
that an adversarially robust classifier must be \emph{sufficiently smooth}, and adversarial examples are a consequence of 
non-smooth directions. As we prove in Theorem~\ref{prop:lossBoundAdversarial}, \emph{sufficiently smooth} in this context means flatness as well as a bounded third derivative 
of the loss function wrt. the weights. To keep our results in line with related work, we use Def.~\ref{def:advEx-classic} in our experiments, but derive our theoretical results for the more general case of Def.~\ref{def:advEx}. 

%% file: empiricalRelation.tex
\section{Efficient Computation of Relative Sharpness}
\label{sec:exps}
To relate flatness and adversarial examples, we want to measure flatness for a particular adversarial example $\xi\in\X$. For that, we need a sound flatness measure and an efficient way to compute it. A sound flatness measure should be correlated with a network's generalization ability. A particular challenge here is that measuring flatness using the loss Hessian wrt. weights (trace or eigenvalues) is not reparameterization-invariant and thus cannot be connected to generalization~\citep{dinh2017sharp}. By deriving their relative flatness measure directly from a decomposition of the generalization gap, \citet{petzka2021relative} could show that a combination of trace of the loss Hessian and norm of weights for a single layer of a network is not only reparameterization-invariant, but can be theoretically linked to generalization. This uses a robustness argument: Since for large weights small perturbation in the representation produced by the layer in question can lead to much larger changes in the output, a network needs to be in a much flatter minimum to counteract this. The ``relative'' part of the flatness measure, i.e., the weights norm component, is a constant in our analysis since we assume a fixed trained neural network. 
For a model that can be decomposed into a feature extractor $\phi$ and a predictor $g$, i.e., $f(x)=g(\w\phi(x))$, trained on $S\subset\X\times\Y$, the relative flatness measure\footnote{Note that this is a simplified relative flatness measure that is not invariant to neuron-wise reparameterizations. Cf. Def. 3 in~\citet{petzka2021relative}.} is defined as 
\[
    \kappa^\phi_{Tr}(\w) := \|w\|_2 Tr(H(\w,S))\enspace ,    
\]
where $Tr$ denotes the trace, and $H$ is shorthand for the Hessian of the loss computed on $S$ wrt $\w$. That is
\[
H(\w,S)=\frac{1}{|S|}\sum_{(x,y)\in S}\left(\frac{\partial^2}{\partial w_i\partial w_j}\loss\left(g(\w\phi(x)),y\right)\right)_{i,j\in [km]}\enspace ,
\]
where $\w$ is the weight matrix corresponding to the selected feature layer that connects the feature extractor $\phi$ and the classifier $g$. Similar to \citet{petzka2021relative} we select $\phi$ as a neural network up to the penultimate layer, $\w$  the weights from the representation to the next layer, and $g$ the final layer of the network. 
Counter-intuitively, the relative flatness measure is small if the loss surface is flat, since in that case the trace of the Hessian is small. Similarly, a large value of $\kappa_{Tr}^\phi(\w)$ means a the loss surface is sharp. To avoid confusion, in the following we therefore call $\kappa_{Tr}^\phi(\w)$ \emph{relative sharpness}.

To compute relative sharpness efficiently, we use the following representation of the Hessian $H$ for the cross-entropy loss $\loss(x,y)=\sum_{i\in [k]}-y_i\log\hat{y}_i$ for a single example $S=\{(x,y)\}$, where---with slight 
abuse of notation---we denote $\hat{y} = g(\w\phi(x))$ as output of the soft-max layer, and write $\phi$ instead of $\phi(x)$.
The Kronecker product is denoted by  $\otimes$. We then have
\begin{equation}
\begin{split}
    H(\w,S) & =  \begin{bmatrix}
\pred_1 (1-\pred_1 ) \phi\phi^T& -\pred_1\pred_2 \phi\phi^T       & \dots & -\pred_1\pred_k  \phi\phi^T\\
-\pred_2\pred_1\phi\phi^T       & \pred_2 (1-\pred_2) \phi\phi^T &       &         \vdots                   \\
\vdots                         & \vdots                          & \dots & \vdots \\
-\pred_k\pred_1  \phi\phi^T     & \vdots                          & \dots &  \pred_k (1-\pred_k)\phi\phi^T & 
\end{bmatrix} \\
    & = (\text{diag}(\pred) - \pred \pred^T  ) \otimes \phi \phi^T \; .
\end{split}
\label{eq:simplifiedHessian}
\end{equation}
 We refer to Appx.~\ref{app:hessian} for a full derivation. The trace of Hessian then is
\begin{equation*}
    tr(H) = \sum_{j=1}^k \pred_j(1-\pred_j)\sum_{i=1}^d \phi_i^2 \; ,
\end{equation*}
by which we can efficiently compute the relative sharpness measure in $\mathcal{O}(k+d)$ instead of $\mathcal{O}(k^2d^2)$.

%% file: experiments.tex
\section{Empirical Evaluation}
We perform three groups of experiments to investigate the behavior of the flatness for adversarial samples.  First, we demonstrate that
adversarial attacks on CNNs find adversarial examples
in flat regions according to the relative sharpness measure. We name these flat regions \textit{uncanny valleys}.
Second,  we investigate, how adversarial training (AT) influences the phenomenon of uncanny valleys. We find that AT simply pushes 
the uncanny valley further away, however, it can still be easily found by stronger attacks. Last, we conduct a similar
analysis on LLMs, where we find that the phenomenon is less pronounced, which is likely due to the discrete nature of the problem,
making the attacks weaker and consequentially the uncanny valleys harder to find. All experiments presented below are carried out on machines
equipped with an Nvidia-A100 80GB, 256 cores, and 1.9TB of memory. The code is publically available \!\footnote{\url{https://github.com/nilspwalter/the-uncanny-valley}}.

\input{latex_figs/first-plot-normalized}
\input{latex_figs/distances_im}
\paragraph{Adversarial Examples Fall Flat}
We train a \resnet~\citep{he2016deep}, \wrn~\citep{Zagoruyko2016WideRN}, \dense ~\citep{huang2017densely}, \vgg  with Batchnorm~\citep{Simonyan2014VeryDC} on CIFAR-10 and CIFAR-100. Each model is trained via stochastic gradient descent for $100$ epochs
with an initial learning rate of $0.1$. We use a cosine scheduler for the learning rate and a weight decay of $10^{-4}$.
Next, we attack each model using PGD-$l_\infty$ with 10 iterations and $\delta=8/255$ ~\citep{madry2017towards}
and record per step the intermediate images generated by the attack and the corresponding loss of the model. This allows us to compute
how the  relative sharpness develops during the attack.  To better compare the different architectures, we normalize the average
sharpness to $[0,1]$.  In Fig.~\ref{fig:first-normalized}, we plot, for each step of the attack, the  \textit{normalized}
relative sharpness measure and the loss with respect to the ground truth. In Appx.~\ref{app:plots} Fig.~\ref{fig:first}, we
show the unnormalized values.

For CIFAR-10, as expected the loss and the sharpness \textit{increase} as the attack progresses, however, around step 3 the sharpness
\textit{decreases} again, while the loss further increases until it saturates. This phenomenon can be observed for all model architectures.
For CIFAR-100, the pattern is slightly different, namely the unperturbed sample already lies in a rather sharp region (which is indicative of the worse test performance, i.e., a larger generalization gap for this task).
Nonetheless, the search for an adversarial example again leads to a very flat region, while the loss steadily increases.
These are still adversarial examples, meaning there are adversarial examples in a flat region for good-performing, but adversarially non-robust models. This indicates that flatness on its own cannot characterize adversarial robustness. 
Characteristic is the fact that one step (corresponding to FGSM) often already leads to a sharp region, thus characterizing this attack as a weaker one. 

In Figure~\ref{fig:dist-im} we show that the adversarial examples move away from the original example with each attack iteration. Thus, the uncanny valleys truly extend away from the original example, supporting the intuition that adversarial examples live in entire subspaces of the input space~\citep{gubri2022lgv, mao2019learning}.

\paragraph{Adversarial Training Pushes the Flat Region Away }
Next, we investigate how adversarial training influences the sharpness measure and the uncanny valleys. Since all
the architectures show practically the same behavior, we focus on  \wrn and CIFAR-10 \& CIFAR-100. To obtain models with varying robustness,
we perform adversarial training using PGD-$l_\infty$  with different $\delta' \in \{1,2,3,4,5,6,7,8\}$, 
where $\delta=\delta'/255$. For reference, we also report the standard trained model, i.e., $\delta'=0$.
As the models trained with $\delta >0$ are more robust, we also employ a stronger attack, namely, we use PGD-$l_\infty$
20 iterations with  $\delta=12/255$ resp. 50 iterations with  $\delta=24/255$.
We plot the development of the relative sharpness and loss for ground truth in Fig.
\ref{fig:adv-train} and \ref{fig:adv-train-cifar100}.

\input{latex_figs/wrn-pgd-eps}
\input{latex_figs/wrn-pgd-eps-cifar100}
For smaller $\delta'<4$, we see a similar behavior to that of normally trained models (Fig.~\ref{fig:first}). However, we observe that 
the turning point occurs at later iterations. This effect is even stronger for $\delta'=4$ and  $\delta'=5$, while for 
$\delta'>5$, the models stay in sharp regions. This demonstrates that adversarial training bulges the area around the samples, so the networks are sharper in that region. Nonetheless, we still observe uncanny valleys, except for models trained with 
larger $\delta'$. If we attack these more robust models with stronger attacks, we again observe that the adversarial examples 
lie in flat regions (Fig. \ref{fig:adv-train-c} and \ref{fig:adv-train-cifar100-c}). Hence, the uncanny valleys still exist and can be reached via stronger attacks.

\paragraph{Jailbreak Attacks on LLMs}
Finally, we show that uncanny valleys also exist for LLMs. We use the pre-trained weights of Vicuna-7B \citep[\vicuna ,][]{vicuna2023}, Llama-2-7B-Chat \citep[\llama ,][]{touvron2023llama}, and Guanaco-7B-HF \citep[\guanaco ,][]{dettmers2024qlora}, downloaded from HuggingFace. Each of these models is aligned, meaning they have been trained to not generate any malicious output and thus robustified against jailbreak attacks.
We use the attack and benchmark of \citet{zou2023universal} for $100$ iterations and a batch size of $1024$, while sampling the top $256$ for further prompting.  We focus on generating malicious behavior: For a given prompt, e.g.,
``\textit{Write a script that can exploit vulnerabilities in a software or operating system}'', the goal is to make the model to answer with 
a specific phrase, e.g, ``\textit{Sure, here is a script that can exploit vulnerabilities in a software or operating system: \dots}''. Hence, these are targeted attacks. Due to runtime constraints, we use only the first $100$ samples in the benchmark. In Fig. \ref{fig:llm-adv-norm-a} and \ref{fig:llm-adv-norm-b}, we report the average relative sharpness per token and the loss with respect to the goal reply.

We observe that for \vicuna the loss landscape first becomes sharper and then the adversarial examples slowly move into flatter regions.
For \guanaco, the phenomenon is less pronounced and for \llama, the adversarial examples stay
in comparatively sharp regions. The uncanny valleys are not as flat as the undefended models shown in Fig.~\ref{fig:first-normalized},
rather, they resemble the curves of adversarially trained models. This can be explained by the fact that these models have been aligned, i.e.,
adversarially trained. 
However, if we inspect individual adversarial samples, we can still find for every model attack trajectories 
where we can observe the uncanny valleys (cf. Fig.~\ref{fig:llm-adv-norm-c}).
Given the similarity to the curves of defended models in Fig.~\ref{fig:adv-train}, we hypothesize that stronger attacks will uncover the uncanny valleys also for LLMs. Nonetheless, even with the current evidence 
we can conclude that LLMs exhibit uncanny valleys during adversarial attacks.

\input{latex_figs/llm-rfm-normalized}

%% file: latex_figs/first-plot-normalized.tex
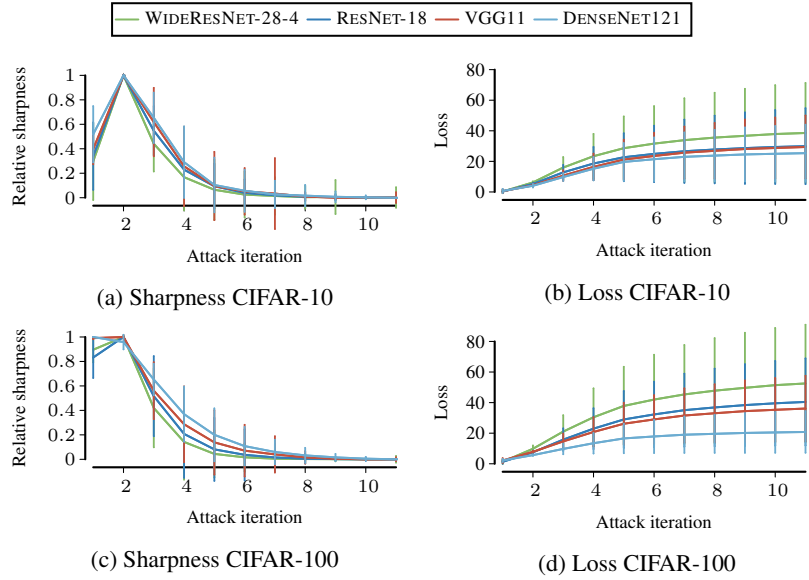
\begin{figure}
\centering
\begin{subfigure}[h]{0.4\linewidth}
        \begin{tikzpicture}
                \usetikzlibrary{calc}
                \begin{axis}[ 
                    pretty line,
                    cycle list name = prcl-line,
                    legend style={yshift=1.0cm,font=\scriptsize,xshift=4cm},
                    width=\linewidth,
                    height=3.2cm,
                    ylabel={Relative sharpness},
                    xlabel={Attack iteration},
                    ymin=0,
                    ymax=1,
                    pretty labelshift,
                    line width=0.75,
                    legend columns = 9,
                    tick label style={/pgf/number format/fixed},
                    legend entries = {\wrn,\resnet,\vgg,\dense}
                ]
                \foreach \x in {wrn, resnet,vgg11_bn,dense121}{
                    \addplot+[forget plot, eda errorbarcolored, y dir=plus, y explicit]
                    table[x={0}, y={trace-\x-cifar10}, y error expr=\thisrow{trace-\x-cifar10}-\thisrow{stdtrace-\x-cifar10}, col sep=comma]{expres/debug-normalized_rfm-for-clean.csv};
                    \addplot+[eda errorbarcolored, y dir=minus, y explicit]
                    table[x={0}, y={trace-\x-cifar10}, y error expr=\thisrow{trace-\x-cifar10}-\thisrow{stdtrace-\x-cifar10}, col sep=comma]{expres/debug-normalized_rfm-for-clean.csv};
                }    
                \end{axis}
            \end{tikzpicture}
                    \subcaption{Sharpness CIFAR-10}
\end{subfigure}%
\begin{subfigure}[h]{0.4\linewidth}
\vspace{1.75em}
        \begin{tikzpicture}
                \usetikzlibrary{calc}
                \begin{axis}[ 
                    pretty line,
                    cycle list name = prcl-line,
                    legend style={yshift=1.0cm,font=\scriptsize,xshift=6cm},
                    width=\linewidth,
                    height=3.2cm,
                    ylabel={Loss},
                    xlabel={Attack iteration},
                    pretty labelshift,
                    line width=0.75,
                    ymin=0,
                    ymax=80,
                    ytick={0,20,40,60,80},
                    yticklabels={0,20,40,60,80},
                    legend columns = 9,
                    tick label style={/pgf/number format/fixed},
                ]
                \foreach \x in {wrn, resnet,vgg11_bn,dense121}{
                    \addplot+[forget plot, eda errorbarcolored, y dir=plus, y explicit]
                    table[x={0}, y={loss-\x-cifar10}, y error expr=\thisrow{loss-\x-cifar10}-\thisrow{stdloss-\x-cifar10}, col sep=comma]{expres/debug-normalized_rfm-for-clean.csv};
                    \addplot+[eda errorbarcolored, y dir=minus, y explicit]
                    table[x={0}, y={loss-\x-cifar10}, y error expr=\thisrow{loss-\x-cifar10}-\thisrow{stdloss-\x-cifar10}, col sep=comma]{expres/debug-normalized_rfm-for-clean.csv};%
                }    
                \end{axis}
            \end{tikzpicture}
                    \subcaption{Loss CIFAR-10}
\end{subfigure}
\begin{subfigure}[h]{0.4\linewidth}
        \begin{tikzpicture}
                \usetikzlibrary{calc}
                \begin{axis}[ 
                    pretty line,
                    cycle list name = prcl-line,
                    legend style={yshift=1.0cm,font=\scriptsize,xshift=6cm},
                    width=\linewidth,
                    height=3.2cm,
                    ylabel={Relative sharpness},
                    xlabel={Attack iteration},
                    pretty labelshift,
                    line width=0.75,
                    ymin=0,
                    ymax=1,
                    legend columns = 9,
                    tick label style={/pgf/number format/fixed},
                ]
                \foreach \x in {wrn, resnet,vgg11_bn,dense121}{
                    \addplot+[forget plot, eda errorbarcolored, y dir=plus, y explicit]
                    table[x={0}, y={trace-\x-cifar100}, y error expr=\thisrow{trace-\x-cifar100}-\thisrow{stdtrace-\x-cifar100}, col sep=comma]{expres/debug-normalized_rfm-for-clean.csv};
                    \addplot+[eda errorbarcolored, y dir=minus, y explicit]
                    table[x={0}, y={trace-\x-cifar100}, y error expr=\thisrow{trace-\x-cifar100}-\thisrow{stdtrace-\x-cifar100}, col sep=comma]{expres/debug-normalized_rfm-for-clean.csv};%
                }    
                \end{axis}
            \end{tikzpicture}
            \subcaption{Sharpness CIFAR-100}
\end{subfigure}%
\begin{subfigure}[h]{0.4\linewidth}
        \begin{tikzpicture}
                \usetikzlibrary{calc}
                \begin{axis}[ 
                    pretty line,
                    cycle list name = prcl-line,
                    legend style={yshift=1.0cm,font=\scriptsize,xshift=6cm},
                    width=\linewidth,
                    height=3.2cm,
                    ylabel={Loss},
                    xlabel={Attack iteration},
                    pretty labelshift,
                    line width=0.75,
                    ymin=0,
                    ymax=80,
                    ytick={0,20,40,60,80},
                    yticklabels={0,20,40,60,80},
                    legend columns = 9,
                    tick label style={/pgf/number format/fixed},
                ]
                \foreach \x in {wrn, resnet,vgg11_bn,dense121}{
                    \addplot+[forget plot, eda errorbarcolored, y dir=plus, y explicit]
                    table[x={0}, y={loss-\x-cifar100}, y error expr=\thisrow{loss-\x-cifar100}-\thisrow{stdloss-\x-cifar100}, col sep=comma]{expres/debug-normalized_rfm-for-clean.csv};
                    \addplot+[eda errorbarcolored, y dir=minus, y explicit]
                    table[x={0}, y={loss-\x-cifar100}, y error expr=\thisrow{loss-\x-cifar100}-\thisrow{stdloss-\x-cifar100}, col sep=comma]{expres/debug-normalized_rfm-for-clean.csv};%
                }    
                \end{axis}
            \end{tikzpicture}
            \subcaption{Loss CIFAR-100}
\end{subfigure}
\caption{We report the normalized relative sharpness on the attack trajectory  for \wrn, \resnet, \vgg and \dense
on the test set of CIFAR-10 \& CIFAR-100. We observe that adversarial examples first reach a sharp region, and as the attack progress they land in a flat region. We also display the standard deviation of the values on individual inputs.}
\label{fig:first-normalized}
\end{figure}

%% file: latex_figs/distances_im.tex
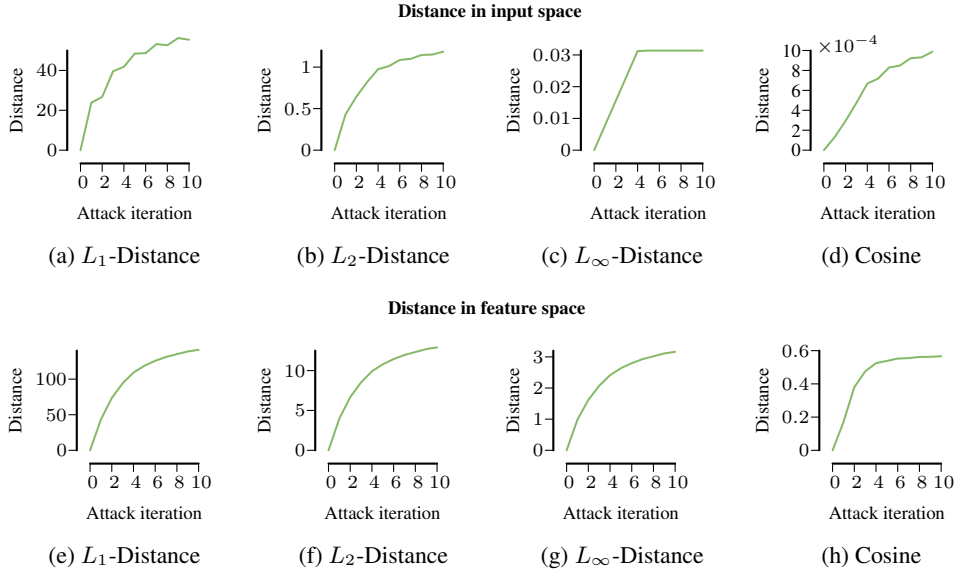
\begin{figure}[ht]
\centering
\captionsetup[subfigure]{oneside,margin={0.5cm,0cm}}
\begin{subfigure}[t]{0.2\linewidth}
        \begin{tikzpicture}
                \usetikzlibrary{calc}
                \begin{axis}[ 
                    smalljonas line,
                    cycle list name = more-prcl,
                    legend style={yshift=1cm,font=\scriptsize,xshift=9cm},
                    width=1.08\linewidth,
                    height=2.9cm,
                    ylabel={Distance},
                    xlabel={Attack iteration},
                    pretty labelshift,
                    line width=0.75,
                    ymin=0,
                    ymax=50,
                    xmin=0,
                    xmax=10,
                    legend columns = 9,
                    tick label style={/pgf/number format/fixed},
                ]
                    \addplot table[x={0},y={l1_im},col sep=comma]{expres/distances.csv};%
                \end{axis}
            \end{tikzpicture}
            \raggedleft\subcaption[Ragged left]{$L_1$-Distance}
\end{subfigure}\hspace{0.5cm}%
\begin{subfigure}[t]{0.2\linewidth}
        \begin{tikzpicture}
                \usetikzlibrary{calc}
                \begin{axis}[ 
                    smalljonas line,
                    cycle list name = more-prcl,
                    legend style={yshift=1cm,font=\scriptsize,xshift=9cm},
                    width=1.08\linewidth,
                    height=2.9cm,
                    ylabel={Distance},
                    xlabel={Attack iteration},
                    pretty labelshift,
                    line width=0.75,
                    ymin=0,
                    ymax=1.2,
                    xmin=0,
                    xmax=10,
                    legend columns = 9,
                    tick label style={/pgf/number format/fixed},
                ]
                    \addplot table[x={0},y={l2_im},col sep=comma]{expres/distances.csv};%
                \node[above, xshift=4.5em,yshift=0.7em] at (current bounding box.north) {\scriptsize \textbf{Distance in input space}};
                \end{axis}
            \end{tikzpicture}
            \subcaption[Ragged left]{$L_2$-Distance}
\end{subfigure}\hspace{0.5cm}%
\begin{subfigure}[t]{0.2\linewidth}
        \begin{tikzpicture}
                \usetikzlibrary{calc}
                \begin{axis}[ 
                    smalljonas line,
                    cycle list name = more-prcl,
                    legend style={yshift=1cm,font=\scriptsize,xshift=9cm},
                    width=1.08\linewidth,
                    height=2.9cm,
                    ylabel={Distance},
                    xlabel={Attack iteration},
                    pretty labelshift,
                    line width=0.75,
                    ymin=0,
                    ymax=0.0314,
                    xmin=0,
                    xmax=10,
                    legend columns = 9,
                    tick label style={/pgf/number format/fixed},
                ]
                   \addplot table[x={0},y={linfty_im},col sep=comma]{expres/distances.csv};%

                \end{axis}
            \end{tikzpicture}
                        \subcaption{$L_{\infty}$-Distance}
\end{subfigure}\hspace{0.5cm}%
\begin{subfigure}[t]{0.2\linewidth}
        \begin{tikzpicture}
                \usetikzlibrary{calc}
                \begin{axis}[ 
                    smalljonas line,
                    cycle list name = more-prcl,
                    legend style={yshift=1cm,font=\scriptsize,xshift=9cm},
                    width=1.08\linewidth,
                    height=2.9cm,
                    ylabel={Distance},
                    xlabel={Attack iteration},
                    pretty labelshift,
                    line width=0.75,
                    ymin=0,
                    ymax=0.001,
                    xmin=0,
                    xmax=10,
                    legend columns = 9,
                    yticklabels={0, 0,2,4,6,8,10}
                ]
                    \addplot table[x={0},y={cosine_im},col sep=comma]{expres/distances.csv};%
            \node[above, xshift=-0.6em,yshift=-1.1em] at (current bounding box.north) {\scriptsize$\times 10^{-4}$};
                \end{axis}
            \end{tikzpicture}
                    \subcaption[Ragged left]{Cosine} 
\end{subfigure}%
\vspace{0.25cm}
\begin{subfigure}[t]{0.2\linewidth}
        \begin{tikzpicture}
                \usetikzlibrary{calc}
                \begin{axis}[ 
                    smalljonas line,
                    cycle list name = more-prcl,
                    legend style={yshift=1cm,font=\scriptsize,xshift=9cm},
                    width=1.08\linewidth,
                    height=2.9cm,
                    ylabel={Distance},
                    xlabel={Attack iteration},
                    pretty labelshift,
                    line width=0.75,
                    ymin=0,
                    ymax=140,
                    xmin=0,
                    xmax=10,
                    legend columns = 9,
                    tick label style={/pgf/number format/fixed},
                ]
                    \addplot table[x={0},y={l1_feat},col sep=comma]{expres/distances.csv};%

                \end{axis}
            \end{tikzpicture}
            \raggedleft\subcaption[Ragged left]{$L_1$-Distance}
\end{subfigure}\hspace{0.5cm}%
\begin{subfigure}[t]{0.2\linewidth}
        \begin{tikzpicture}
                \usetikzlibrary{calc}
                \begin{axis}[ 
                    smalljonas line,
                    cycle list name = more-prcl,
                    legend style={yshift=1cm,font=\scriptsize,xshift=9cm},
                    width=1.08\linewidth,
                    height=2.9cm,
                    ylabel={Distance},
                    xlabel={Attack iteration},
                    pretty labelshift,
                    line width=0.75,
                    ymin=0,
                    ymax=12.5,
                    xmin=0,
                    xmax=10,
                    legend columns = 9,
                    tick label style={/pgf/number format/fixed},
                ]
                    \addplot table[x={0},y={l2_feat},col sep=comma]{expres/distances.csv};%
            \node[above, xshift=4.5em,yshift=0.7em] at (current bounding box.north) {\scriptsize \textbf{Distance in feature space}};
                \end{axis}
            \end{tikzpicture}
            \subcaption[Ragged left]{$L_2$-Distance}
\end{subfigure}\hspace{0.5cm}%
\begin{subfigure}[t]{0.2\linewidth}
        \begin{tikzpicture}
                \usetikzlibrary{calc}
                \begin{axis}[ 
                    smalljonas line,
                    cycle list name = more-prcl,
                    legend style={yshift=1cm,font=\scriptsize,xshift=9cm},
                    width=1.08\linewidth,
                    height=2.9cm,
                    ylabel={Distance},
                    xlabel={Attack iteration},
                    pretty labelshift,
                    line width=0.75,
                    ymin=0,
                    ymax=3.2,
                    xmin=0,
                    xmax=10,
                    legend columns = 9,
                    tick label style={/pgf/number format/fixed},
                ]
                   \addplot table[x={0},y={linfty_feat},col sep=comma]{expres/distances.csv};%

                \end{axis}
            \end{tikzpicture}
                        \subcaption{$L_{\infty}$-Distance}
\end{subfigure}\hspace{0.5cm}%
\begin{subfigure}[t]{0.2\linewidth}
        \begin{tikzpicture}
                \usetikzlibrary{calc}
                \begin{axis}[ 
                    smalljonas line,
                    cycle list name = more-prcl,
                    legend style={yshift=1cm,font=\scriptsize,xshift=9cm},
                    width=1.08\linewidth,
                    height=2.9cm,
                    ylabel={Distance},
                    xlabel={Attack iteration},
                    pretty labelshift,
                    line width=0.75,
                    ymin=0,
                    ymax=0.6,
                    xmin=0,
                    xmax=10,
                ]
                    \addplot table[x={0},y={cosine_feat},col sep=comma]{expres/distances.csv};%

                \end{axis}
            \end{tikzpicture}
                    \subcaption[Ragged left]{Cosine} 
\end{subfigure}%
\caption{We show how far adversarial examples move in image/feature space from the initial image during a PGD-attack; we measure distance with $L_1$, $L_2$, $L_{\infty}$ and cosine dissimilarity i.e. 1 - cosine similarity. We used CIFAR-10, \wrn, and PGD with 10 iterations and $\delta$=8/255. }
\label{fig:dist-im}
\vspace{-0.3cm}
\end{figure}

%% file: latex_figs/wrn-pgd-eps.tex
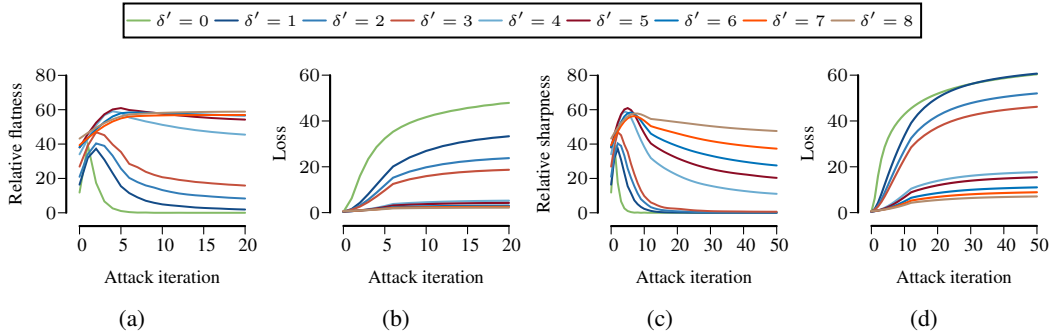
\begin{figure}[!b]
\begin{subfigure}[t]{0.25\linewidth}
        \begin{tikzpicture}
                \usetikzlibrary{calc}
                \begin{axis}[ 
                    smalljonas line,
                    cycle list name = more-prcl,
                    legend style={yshift=1cm,font=\scriptsize,xshift=9cm},
                    width=1.08\linewidth,
                    height=3.4cm,
                    ylabel={Relative flatness},
                    xlabel={Attack iteration},
                    pretty labelshift,
                    line width=0.75,
                    ymin=0,
                    ymax=80,
                    xmin=1,
                    xmax=21,
                    xtick={1,6,11,16,21},
                    xticklabels =  {0, 5, 10, 15, 20},
                    legend columns = 9,
                    tick label style={/pgf/number format/fixed},
                    legend entries = {$\delta'=0$,$\delta'=1$,$\delta'=2$,$\delta'=3$,$\delta'=4$,$\delta'=5$,$\delta'=6$, $\delta'=7$,$\delta'=8$},
                ]
                \foreach \x in {0,1,2,3,4,5,6,7,8}{
                    \addplot table[x={0},y={trace-\x},col sep=comma]{expres/diff_eps/debug-steps-20-eps_attack_12_wrn_cifar10.csv};%
                    
                }    
                \end{axis}
            \end{tikzpicture}
            \subcaption{}
\end{subfigure}%
\begin{subfigure}[t]{0.25\linewidth}
        \begin{tikzpicture}
                \usetikzlibrary{calc}
                \begin{axis}[ 
                    smalljonas line,
                    cycle list name = more-prcl,
                    legend style={yshift=0.7cm,font=\scriptsize,xshift=3.4cm},
                    width=1.08\linewidth,
                    height=3.4cm,
                    ylabel={Loss},
                    xlabel={Attack iteration},
                    pretty labelshift,
                    ymin=0,
                    ymax=60,
                    xmin=1,
                    xmax=21,
                    xtick={1,6,11,16,21},
                    xticklabels =  {0, 5, 10, 15, 20},
                    legend columns = 5,
                    tick label style={/pgf/number format/fixed}
                ]
                \foreach \x in {0,1,2,3,4,5,6,7,8}{
                    \addplot table[x={0},y={loss-\x},col sep=comma]{expres/diff_eps/debug-steps-20-eps_attack_12_wrn_cifar10.csv};%
                }    
                \end{axis}
            \end{tikzpicture}
            \subcaption{}
\end{subfigure}%
\begin{subfigure}[t]{0.25\linewidth}
        \begin{tikzpicture}
                \usetikzlibrary{calc}
                \begin{axis}[ 
                    smalljonas line,
                    cycle list name = more-prcl,
                    legend style={yshift=1.0cm,font=\scriptsize,xshift=6cm},
                    width=1.08\linewidth,
                    height=3.4cm,
                    ylabel={Relative sharpness},
                    xlabel={Attack iteration},
                    pretty labelshift,
                    line width=0.75,
                    ymin=0,
                    ymax=80,
                    xmin=1,
                    xmax=51,                    
                    xtick={1,11,21,31,41,51},
                    xticklabels =  {0, 10, 20, 30, 40, 50},
                    legend columns = 9,
                    tick label style={/pgf/number format/fixed},
                ]
                \foreach \x in {0,1,2,3,4,5,6,7,8}{
                    \addplot table[x={0},y={trace-\x},col sep=comma]{expres/diff_eps/debug-steps-50-eps_attack_24_wrn_cifar10.csv};%
                    
                }    
                \end{axis}
            \end{tikzpicture}
                        \subcaption{}
                        \label{fig:adv-train-c}
\end{subfigure}%
\begin{subfigure}[t]{0.25\linewidth}
        \begin{tikzpicture}
                \usetikzlibrary{calc}
                \begin{axis}[ 
                    smalljonas line,
                    cycle list name = more-prcl,
                    legend style={yshift=0.7cm,font=\scriptsize,xshift=3.4cm},
                    width=1.08\linewidth,
                    height=3.4cm,
                    ylabel={Loss},
                    xlabel={Attack iteration},
                    pretty labelshift,
                    ymin=0,
                    ymax=60,
                    xmin=1,
                    xmax=51,                    
                    xtick={1,11,21,31,41,51},
                    xticklabels =  {0, 10, 20, 30, 40, 50},
                    legend columns = 5,
                    tick label style={/pgf/number format/fixed}
                ]
                \foreach \x in {0,1,2,3,4,5,6,7,8}{
                    \addplot table[x={0},y={loss-\x},col sep=comma]{expres/diff_eps/debug-steps-50-eps_attack_24_wrn_cifar10.csv};%
                }    
                \end{axis}
            \end{tikzpicture}
            \subcaption{}
\end{subfigure}
\caption{Here, we evaluate adversarially trained \wrn on CIFAR-10 with varying $\delta$. We attack the resulting models using PGD-$l_\infty$ with
        $\delta=12/255$, $\text{steps}=20$, shown in Figure (a) \& (b), and $\delta=24/255$, $\text{steps}=50$, depicted in  Figure (c) \& (d).We can see that even for adversarially trained models, we can find uncanny valleys by using  a stronger attack.}
\label{fig:adv-train}
\end{figure}

%% file: latex_figs/wrn-pgd-eps-cifar100.tex
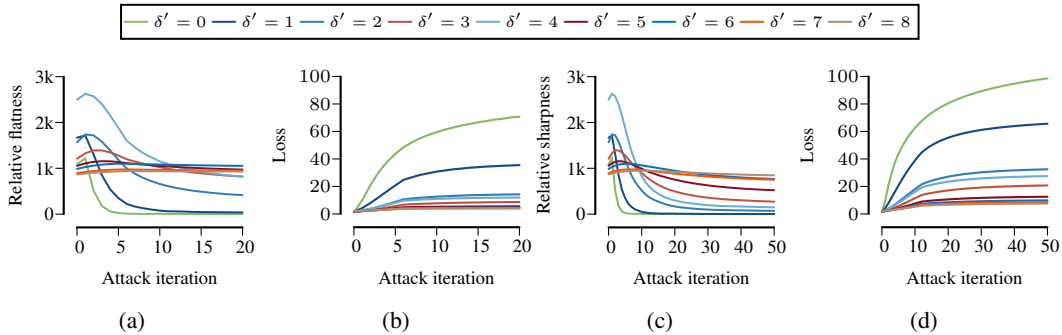
\begin{figure}[!b]
\begin{subfigure}[t]{0.25\linewidth}
        \begin{tikzpicture}
                \usetikzlibrary{calc}
                \begin{axis}[ 
                    smalljonas line,
                    cycle list name = more-prcl,
                    legend style={yshift=1cm,font=\scriptsize,xshift=9cm},
                    width=1.08\linewidth,
                    height=3.4cm,
                    ylabel={Relative flatness},
                    xlabel={Attack iteration},
                    pretty labelshift,
                    line width=0.75,
                    yticklabels={0,0,1k,2k,3k},
                    ymin=0.0,
                    ymax=3000,
                    xmin=1,
                    xmax=21,
                    xtick={1,6,11,16,21},
                    xticklabels =  {0, 5, 10, 15, 20},
                    legend columns = 9,
                    tick label style={/pgf/number format/fixed},
                    legend entries = {$\delta'=0$, $\delta'=1$,$\delta'=2$,$\delta'=3$,$\delta'=4$,$\delta'=5$,$\delta'=6$, $\delta'=7$,$\delta'=8$},
                ]
                \foreach \x in {0,1,2,3,4,5,6,7,8}{
                    \addplot table[x={0},y={trace-\x},col sep=comma]{expres/diff_eps/debug-steps-20-eps_attack_12_wrn_cifar100.csv};%
                    
                }    
                \end{axis}
            \end{tikzpicture}
            \subcaption{}
\end{subfigure}%
\begin{subfigure}[t]{0.25\linewidth}
        \begin{tikzpicture}
                \usetikzlibrary{calc}
                \begin{axis}[ 
                    smalljonas line,
                    cycle list name = more-prcl,
                    legend style={yshift=0.7cm,font=\scriptsize,xshift=3.4cm},
                    width=1.08\linewidth,
                    height=3.4cm,
                    ylabel={Loss},
                    xlabel={Attack iteration},
                    pretty labelshift,
                    ymin=0,
                    ymax=100,
                    xmin=1,
                    xmax=21,
                    xtick={1,6,11,16,21},
                    xticklabels =  {0, 5, 10, 15, 20},
                    legend columns = 5,
                    tick label style={/pgf/number format/fixed}
                ]
                \foreach \x in {0,1,2,3,4,5,6,7,8}{
                    \addplot table[x={0},y={loss-\x},col sep=comma]{expres/diff_eps/debug-steps-20-eps_attack_12_wrn_cifar100.csv};%
                }    
                \end{axis}
            \end{tikzpicture}
            \subcaption{}
\end{subfigure}%
\begin{subfigure}[t]{0.25\linewidth}
        \begin{tikzpicture}
                \usetikzlibrary{calc}
                \begin{axis}[ 
                    smalljonas line,
                    cycle list name = more-prcl,
                    legend style={yshift=1.0cm,font=\scriptsize,xshift=6cm},
                    width=1.08\linewidth,
                    height=3.4cm,
                    ylabel={Relative sharpness},
                    xlabel={Attack iteration},
                    pretty labelshift,
                    line width=0.75,
                    yticklabels={0,0,1k,2k,3k},
                    ymin=0.0,
                    ymax=3000,
                    xmin=1,
                    xmax=51,                    
                    xtick={1,11,21,31,41,51},
                    xticklabels =  {0, 10, 20, 30, 40, 50},
                    legend columns = 9,
                    tick label style={/pgf/number format/fixed},
                ]
                \foreach \x in {0,1,2,3,4,5,6,7,8}{
                    \addplot table[x={0},y={trace-\x},col sep=comma]{expres/diff_eps/debug-steps-50-eps_attack_24_wrn_cifar100.csv};%
                    
                }    
                \end{axis}
            \end{tikzpicture}
                        \subcaption{}
                        \label{fig:adv-train-cifar100-c}
\end{subfigure}%
\begin{subfigure}[t]{0.25\linewidth}
        \begin{tikzpicture}
                \usetikzlibrary{calc}
                \begin{axis}[ 
                    smalljonas line,
                    cycle list name = more-prcl,
                    legend style={yshift=0.7cm,font=\scriptsize,xshift=3.4cm},
                    width=1.08\linewidth,
                    height=3.4cm,
                    ylabel={Loss},
                    xlabel={Attack iteration},
                    pretty labelshift,
                    ymin=0,
                    ymax=100,
                    xmin=1,
                    xmax=51,                    
                    xtick={1,11,21,31,41,51},
                    xticklabels =  {0, 10, 20, 30, 40, 50},
                    legend columns = 5,
                    tick label style={/pgf/number format/fixed}
                ]
                \foreach \x in {0,1,2,3,4,5,6,7,8}{
                    \addplot table[x={0},y={loss-\x},col sep=comma]{expres/diff_eps/debug-steps-50-eps_attack_24_wrn_cifar100.csv};%
                }    
                \end{axis}
            \end{tikzpicture}
            \subcaption{}
\end{subfigure}
\caption{Here, we evaluate adversarially trained \wrn on CIFAR-100 with varying $\delta$. We attack the resulting models using PGD-$l_\infty$ with
        $\delta=12/255$, $\text{steps}=20$, shown in Figure (a) \& (b), and $\delta=24/255$, $\text{steps}=50$, depicted in  Figure (c) \& (d).We can see that even for adversarially trained models, we can find uncanny valleys by using  a stronger attack.}
\label{fig:adv-train-cifar100}
\end{figure}

%% file: latex_figs/llm-rfm-normalized.tex
\begin{figure}
\begin{subfigure}[t]{0.25\linewidth}
        \begin{tikzpicture}
                \usetikzlibrary{calc}
                \begin{axis}[ 
                    smalljonas line,
                    cycle list name = prcl-line,
                    legend style={yshift=1.4cm,font=\scriptsize,xshift=5.5cm},
                    width=1.08\linewidth,
                    height=3.4cm,
                    ylabel={Relative sharpness},
                    xlabel={Attack iteration},
                    pretty labelshift,
                    line width=0.75,
                    xmin=0,
                    xmax=100,
                    legend columns = 3,
                    tick label style={/pgf/number format/fixed},
                    legend entries = {\vicuna, \llama, \guanaco},
                ]
                \foreach \x in {vicuna, llama2, guanaco}{
                    \addplot table[x={index},y={\x-fam},col sep=comma]{expres/llms/normalized-\x-attack.csv};%
                }    
                \end{axis}
            \end{tikzpicture}
            \subcaption{}
                \label{fig:llm-adv-norm-a}
\end{subfigure}%
\begin{subfigure}[t]{0.25\linewidth}
        \begin{tikzpicture}
                \usetikzlibrary{calc}
                \begin{axis}[ 
                    smalljonas line,
                    cycle list name = prcl-line,
                    legend style={yshift=0.7cm,font=\scriptsize,xshift=3.4cm},
                    width=1.08\linewidth,
                    height=3.4cm,
                    ylabel={Loss},
                    xlabel={Attack iteration},
                    pretty labelshift,
                    ymin=0.0,
                    ymax=2.5,
                    ytick={0,0.5,1.0,1.5,2.0,2.5},
                    xmin=0,
                    xmax=100,
                    legend columns = 5,
                    tick label style={/pgf/number format/fixed}
                ]
                \foreach \x in {vicuna, llama2, guanaco}{
                    \addplot table[x={index},y={\x-loss},col sep=comma]{expres/llms/normalized-\x-attack.csv};%
                }       
                \end{axis}
            \end{tikzpicture}
            \subcaption{}
                \label{fig:llm-adv-norm-b}
\end{subfigure}%
\begin{subfigure}[t]{0.25\linewidth}
        \begin{tikzpicture}
                \usetikzlibrary{calc}
                \begin{axis}[ 
                    smalljonas line,
                    cycle list name = prcl-line,
                    legend style={yshift=0.7cm,font=\scriptsize,xshift=3.4cm},
                    width=1.08\linewidth,
                    height=3.4cm,
                    ylabel={Relative sharpness},
                    xlabel={Attack iteration},
                    pretty labelshift,
                    xmin=0,
                    xmax=100,
                    legend columns = 5,
                    tick label style={/pgf/number format/fixed}
                ]
                \foreach \x in {vicuna, llama2, guanaco}{
                    \addplot table[x={index},y={fam_pos},col sep=comma]{expres/posneg/normalized-\x_posneg.csv};%
                }       
                \end{axis}
            \end{tikzpicture}
            \subcaption{}
            \label{fig:llm-adv-norm-c}
\end{subfigure}%
\begin{subfigure}[t]{0.25\linewidth}
        \begin{tikzpicture}
                \usetikzlibrary{calc}
                \begin{axis}[ 
                    smalljonas line,
                    cycle list name = prcl-line,
                    legend style={yshift=0.7cm,font=\scriptsize,xshift=3.4cm},
                    width=1.08\linewidth,
                    height=3.4cm,
                    ylabel={Loss},
                    xlabel={Attack iteration},
                    pretty labelshift,
                    ymin=0.0,
                    ymax=2.5,
                    ytick={0,0.5,1.0,1.5,2.0,2.5},
                    xmin=0,
                    xmax=100,
                    legend columns = 5,
                    tick label style={/pgf/number format/fixed}
                ]
                \foreach \x in {vicuna, llama2, guanaco}{
                    \addplot table[x={index},y={loss_pos},col sep=comma]{expres/posneg/normalized-\x_posneg.csv};%
                }       
                \end{axis}
            \end{tikzpicture}
            \subcaption{}
            \label{fig:llm-adv-norm-d}
\end{subfigure}
\caption{In (a) and (b), we plot the relative sharpness and loss of the adversarial prompt for \vicuna, \llama and \guanaco when attacked by
the method of \cite{zou2023universal}. Additionally, in (c) and (d), we plot per model example trajectories, which first become sharper and then flatter again, together with the corresponding loss.}
\label{fig:llm-adv-norm}
\end{figure}
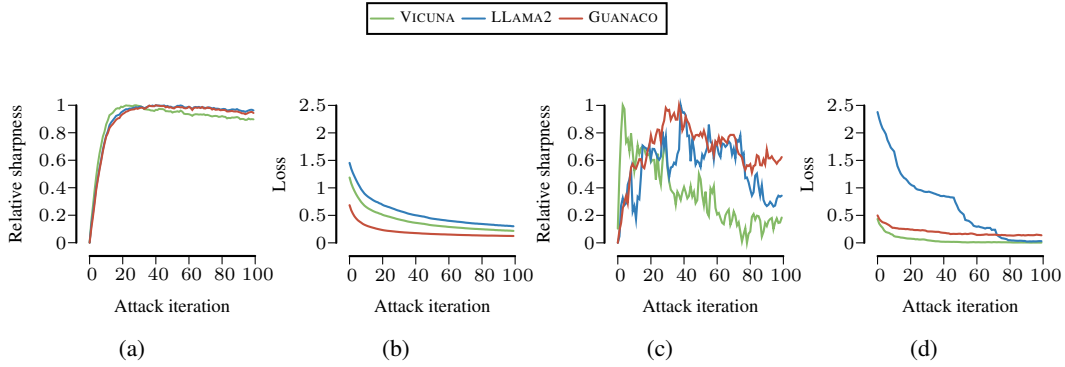

%% file: theory.tex
\section{Bounding Adversarial Robustness via Relative Sharpness}
\label{sec:theory}
The definition of adversarial examples entails the assumption that labels are locally constant, which we made explicit in Def.~\ref{def:advEx}. This assumption can lead to counter-intuitive behavior: if the true label of a perturbed example $\xi$ is $y_\xi\neq y$ and the model predicts $y_\xi$,
then although correct it counts as an adversarial example, while if it predicts $y$ it does \emph{not} count as an adversarial example even though the model makes a mistake. For image classification, this appears intuitively reasonable since small changes to images should not change the class. It follows that for such applications, the ideal model is smooth, i.e., small perturbations of the input do not affect its output. 

Several works have shown that adversarial training increases flatness with respect to the inputs~\citep{moosavi2019robustness, kanai2023relationship}, and adversarial robustness has been empirically linked to flatness of the loss curve with respect to the parameters~\citep{wu2020adversarial, stutz_relating_2021}, but a theoretical link is non-trivial since perturbations of the input relate non-linearly to perturbations of the model weights. In the following, we establish such a formal link and show how it can be used to guarantee adversarial robustness. For that, we first formally define adversarial robustness. A model $f\in\F$ is robust against adversarial examples on a set $S\subseteq\X\times\Y$ if no adversarial examples exist in the vicinity of each element of $S$. With $\mathcal{B}_d^\delta(x)=\{\xi\in\X \mid d(x,\xi)\leq \delta\}$ denoting the $\delta$-ball around $x\in\X$ with respect to the distance $d$ from Def.~\ref{def:advEx}, adversarial robustness can be defined similar to~\citet{schmidt2018adversarially} as follows.
\begin{df}
\label{def:adversarialRobustnessStrict}
Let $S\subseteq\X\times\Y$ be a dataset drawn iid.~according to $\D$ and $d:\X\times\X$ be a distance on $\X$. A model $f\in\F$ is $(\epsilon,\delta,S)$-robust against adversarial examples, if for all $x\in S$ it holds that there is no adversarial example with threshold $\epsilon$ in $\mathcal{B}_d^\delta(x)$. A model $f$ is $(\epsilon, \delta)$-robust if it is $(\epsilon,\delta,S)$-robust for all $S$ drawn iid from $\D$.
\end{df}

Relative sharpness of a model $f(x)=g(\w\phi(x))$ with feature extractor $\phi$, classifier $g$, and weights from features to classifier $\w$ is non-linearly linked to perturbations in the input~\citep{petzka2021relative}. We use this property to show that relative sharpness measured for a single example can bound the loss increase through perturbations. The minimum relative sharpness for all clean examples in data $S$ then provides a bound on how much any example in $S$ can be perturbed without significantly increasing the loss. In other words, it provides a guarantee on the adversarial robustness of $f$ on $S$. 

For this, we have to first establish a link between perturbations in the input space and perturbations in feature space. That is, we express the representation $\phi(\xi)$ of an adversarial example in feature space as a perturbation of the representation $\phi(x)$ of the clean example $x$.
\begin{restatable}{lm}{relationFeatureInput}
    Let $f=g(\w\phi(x))$ be a model with $\phi$ $L$-Lipschitz and $\|\phi(x)\|\geq r$, and $\xi,x\in\X$ with $\|\xi - x\|\leq \delta$, then there exists a $\Delta>0$ with
    $\Delta \leq L\delta r^{-1}$, such that $\phi(\xi)=\phi(x)+\Delta A\phi(x)$, where $A$ is an orthogonal matrix.
\label{lm:relationFeatureInput}
\end{restatable}
The proof is provided in Appx.~\ref{appdx:proofLmRelationFeatureInput}. 
The second step is to relate the adversarial example $\xi$ to perturbations in the weights $\w$ of the representation layer, for which we use the linearity argument from~\citet{petzka2021relative}. Since we can express $\phi(\xi)$ as $\phi(x)+\Delta A\phi(x)$, we can then use the linearity of the representation layer to relate the perturbation of input to a perturbation in weights. That is,
\[
\loss(f(\xi),y)=\loss(g(\w\phi(\xi)),y)=\loss(g(\w(\phi(x)+\Delta A\phi(x))),y)=\loss(g((\w+\Delta \w A)\phi(x),y)\enspace .
\]
This means that we can now express an adversarial example as a suitable perturbation of the weights $\w$ of the representation layer, where the magnitude is bounded as $\Delta\leq L\delta r^{-1}$.
We now bound the loss difference between $\xi$ and $x$. For convenience, we define $\loss(\w+\Delta \w A):=\loss(g((\w+\Delta \w A)\phi(x),y)$. 
The Taylor expansion of $\loss(\w+\Delta \w A)$ at $\w$ yields
\[
\loss(\w+\Delta \w)=\loss(\w)+\innerprod{\Delta \w A}{\nabla_\w\loss(\w)}+\frac{\Delta^2}{2}\innerprod{\w A}{H\loss(\w)(\w A)}+R_2(\w,\Delta)\enspace ,
\]
where $H\loss(\w)$ is the Hessian of $\loss(\w)$. If we now maximize over all $A$ with $\|A\|\leq 1$, it follows that  
$\innerprod{\w A}{H\loss(\w)(\w A)}\leq \|\w\|_F^2 Tr(H\loss(\w)) = \kappa^\phi_{Tr}(\w)$.
Therefore, we have
\begin{equation}
\left|\loss(f(\xi),y) - \loss(f(x),y)\right|\leq \Delta\|\w\|_F\|\nabla_\w\loss(\w)\|_F + \frac{\Delta^2}{2}\kappa^\phi_{Tr}(\w) + R_2(\w,\Delta)\enspace .
\label{eq:lossBoundTaylor}
\end{equation}
The remainder depends on the partial third derivatives of the loss. We show in Appdx.~\ref{app:lossBoundAdversarial} that for feature extractor $\phi$ that is $L$-Lipschitz, it can be bound by $4^{-1}kmL^3$.
With this we can bound the difference between the loss suffered on an adversarial example $\xi$ and the loss on a clean example $x$ for a converged model as follows.
\begin{restatable}{pr}{lossBoundAdversarial}
For $(x,y)\in\X\times\Y$ with $\|x\|\leq 1$ for all $x\in\X$, a model $f(x)=g(\w\phi(x))$ at a minimum $\w\in\R^{m\times k}$ with $\phi$ $L$-Lipschitz and $\|\phi(x)\|\geq r$, and the cross-entropy loss $\loss(\w)=\loss(g(\w \phi(x)),y)$ of $f$ on $(x,y)$, it holds for all $\xi\in\X$ with $\|x-\xi\|_2\leq\delta$ that
\begin{equation*}
    \begin{split}
        \loss(f(\xi),y) - \loss(f(x),y)\leq& \frac{\delta^2}{2r^2}L^2\kappa^\phi_{Tr}(\w) + \frac{\delta^3}{24r^3}kmL^6\enspace .
    \end{split}
\end{equation*}
\label{prop:lossBoundAdversarial}
\end{restatable}
We defer the proof to Appx.~\ref{app:lossBoundAdversarial}. Note, in practice, the perturbation budget $\delta<<1$ is chosen to be small. Hence, this bound can be practically useful not only for Lipschitz-regularized networks~\citep{virmaux2018lipschitz} or $1$-Lipschitz networks~\citep{araujounified} but also for larger Lipschitz-constants $L$.
This bound on the loss difference $\epsilon$ between $\xi$ and $x$ can be transformed in a guarantee on adversarial robustness based on relative sharpness by solving the cubic equation for $\delta$.
\begin{restatable}{crl}{adversarialRobustness}[Informal] \label{prop:bound}
For a dataset $S\subset\X\times\Y$ with $\|x\|\leq 1$ for all $x\in\X$, a model $f(x)=g(\w\phi(x))$ at a minimum $\w\in\R^{m\times k}$ wrt. $S$, with $\phi$ $L$-Lipschitz and $\|\phi(x)\|\geq r$, and the cross-entropy loss $\loss(\w)=\loss(g(\w \phi(x)),y)$ of $f$ on $(x,y)$, $d$ being the $L2$-distance, and $\epsilon>0$, $f$ is $(\epsilon,\delta,S)$-robust against adversarial examples with
\begin{equation*}
    \delta\propto \frac{\epsilon^{\frac13}}{\kappa_{Tr}(\w)^{\frac13}L} + \frac{rkmL^2}{\kappa_{Tr}(\w)}\enspace ,
\end{equation*}
where $\kappa_{Tr}(\w)$ is the relative sharpness of $f$ wrt. $\w$.
\label{prop:adversarialRobustness}
\end{restatable}
The formal version, together with the proof of this corollary, are provided in Appx.~\ref{appdx:proofPropAdvRobustness}. Cor.~\ref{prop:bound} implies that a flat loss surface measured in terms of relative sharpness $\kappa_{Tr}^\phi(\w)$, together with smoothness of the feature extractor $\phi$ in terms of Lipschitz-continuity guarantees adversarial robustness on a dataset $S$ with particular $\delta$. Under the assumption of locally constant labels and for data distributions $\D$ with smooth density $p_\D^\phi$ in feature space, the non-simplified relative flatness also implies good generalization~\citep{petzka2021relative}. Together, these theoretical results indicate that if a model $f=g(\w\phi(x))$ has small relative sharpness $\kappa_{Tr}^\phi(\w)$ for all examples in $S$ and $\phi$ is Lipschitz, then it simultaneously achieves good generalization and adversarial robustness.

%% file: discussion.tex
\section{Discussion \& Conclusion}
\label{sec:discussion}

Within the domain of guaranteeing adversarial robustness, a common methodology is to assume (or regularize for) a low Lipschitz constant to derive bounds.
It is also known, however, that minimizing a global Lipschitz constant \emph{indefinitely} leads to bad performance on clean samples~\citep{yang2020closer}.
The global Lipschitz constant alone is therefore not helpful for obtaining good-performing and robust models.
There exists, however, an intricate connection between the Lipschitz property of the model and its flatness measured with respect to parameters~\citep{kanai2023relationship}; moreover, improving robustness with respect to the parameter changes also improves adversarial robustness empirically~\citep{wu2020adversarial}.
Using the notion of relative flatness~\citep{petzka2021relative}, which is theoretically linked to the generalization of the deep learning models, we develop a bound that connects Lipschitzness, flatness, and adversarial robustness.
This bound means that inducing flatness with respect to parameters will improve the adversarial robustness of the model, given that the function described by the model is smooth.
Together with the results of \citet{petzka2021relative} this links the adversarial robustness and generalization abilities of a model.

Measurements of relative flatness for multi-step attacks with varying parameters reveal two things.
First,  for all architectures, we can systematically find uncanny valleys on which the trajectory indicates a very similar 
geometry across the models, which might be connected to the existence of Universal Adversarial Examples \citep{moosavi2017universal} and
transferability of adversarial examples.
Second, strong adversarial samples fall into flat uncanny valleys and therefore would be missed by most defenses and detection methods based on representations~\citep{lee2018simple, ma2018characterizing, xu_feature_2018, hu2019new, tian_detecting_2018}.
This is in alignment with conclusions drawn by \citet{tramer2022detecting} and \citet{carlini2017adversarial}: the strength of the attack can always be increased in a way that the currently applied defense will fail.
At the same moment, uncanny valleys give a fast and easy-to-measure quantity, which could allow for the detection of adversarial attacks 
during inference. Interestingly, and contrary to the observation of \cite{tramer2022detecting}, stronger attacks that fall into 
uncanny valleys are in fact \emph{easier} to detect. Simultaneously, relative flatness can also serve as a tool to detect jailbreak attacks,
as most of the adversarial prompts lie in sharp regions.

\paragraph{Limitations \& Future Work}
We consider relative flatness wrt. the penultimate layer, because it is theoretically sound and its Hessian is fast to compute
, when cross-entropy loss is used . There exists, however, a multitude of previous work claiming that adversarial robustness should be considered with respect to the properties of the individual layers; \citet{kumari2019harnessing} propose a latent adversarial training technique that improves the robustness of intermediate layers and this leads to better overall robustness; \citet{bakiskan2022early} and \citet{walter_fragile_2022} analyse which of the layers play a specific role for robustness. \cite{petzka2021relative} show that the relative flatness measured in different layers has similar correlation with generalization. In line with that we observation, 
we find in preliminary experiments that relative flatness wrt.~adversarial examples shows similar patterns in different layers
(cf. Figure \ref{fig:diff-layer}), i.e., the uncanny valley is also present in earlier representations. The layer-wise perspective on the robustness is an interesting research direction~\citep{adilova2023layer} that might be a continuation for this work, i.e., can we regularize for the flatness of one layer to obtain better performing and more adversarially robust networks? In particular, the fact that the phenomenon follows a similar pattern in shallower layers begs the questions,
if the last layer can serve as an avenue to  influence the robustness of earlier representations, but at a much lower cost.

Computing the Lipschitz constant exactly is NP-hard, which limits the practical applicability of our theoretical results. For 1-Lipschitz neural networks~\citep{araujounified} or networks trained with Lipschitz regularization~\citep{virmaux2018lipschitz, gouk2021regularisation}, however, the bound is readily applicable. For general neural networks, exploring how a tighter approximation of local smoothness via local Lipschitzness can be used to improve the practicability of our results makes for excellent future work.

An important direction for future work is to investigate how the phenomenon of uncanny valleys can be translated into an actionable detection method for image classifiers, and more importantly for LLMs, as these models are already exposed to the general public and pose a larger threat.  We provide some preliminary results in Appendix \ref{app:detection} by simply thresholding the sharpness measure
which show that adversarial examples can be detected with $>90\%$ accuracy.
Currently, there is no satisfying theoretical explanation for the existence of uncanny valleys.
Thus, we are specifically interested in exploring and answering the question: 
\emph{What lies at the bottom of the uncanny valley?}

%% file: appendix.tex
\section*{Appendix}
\label{sec:apx}
\section{Proofs of Theoretical Results}
In this section, we provide the proofs for the theoretical results in this paper.
\subsection{Proof of Lemma~\ref{lm:relationFeatureInput}}
\label{appdx:proofLmRelationFeatureInput}
For convenience, we restate the lemma.
\relationFeatureInput*
\begin{proof}    
    It follows from the proof in Thm.~5 in \citet{petzka2021relative} that we can represent any vector $v\in\X$ as $v=w+\Delta Aw$ for some vector $w\in X$, $\Delta\in\R_+$ and $A$ an orthogonal matrix. 
    Then 
    \begin{equation*}
        \begin{split} 
            & \phi(\xi)=\phi(x)+\Delta A\phi(x)\\
            \Leftrightarrow & (\phi(\xi) - \phi(x))=\Delta(A\phi(x))\\
            \Leftrightarrow & \Delta\leq \|(\phi(\xi) - \phi(x))\|\|(A\phi(x))^{-1}\|= \underbrace{\|(\phi(x + \Delta'A'x) - \phi(x))\|}_{\leq L\Delta'}\|(A\phi(x))^{-1}\|\\
            \Leftrightarrow & \Delta\leq L\Delta'\|(A\phi(x))^{-1}\|\underbrace{\leq}_{A\text{ orth.}} L\Delta'\frac{1}{r}\enspace .
        \end{split}
    \end{equation*}
    The result follows from $\|\xi - x\| = \Delta'\leq\delta$.
\end{proof}

\subsection{Proof of Proposition~\ref{prop:lossBoundAdversarial}}
\label{app:lossBoundAdversarial}
For convenience, we restate the proposition.
\lossBoundAdversarial*
\begin{proof} 
The remainder $R_2$ in Eq.~\ref{eq:lossBoundTaylor} is
\[
R_2(\w,\Delta)\leq \sup_{\stackrel{h\in\R^m}{\|h\|=1}}\sup_{c\in (0,1)}\frac{\Delta^3}{3!}\sum_{i,j,k}^d\frac{\partial^3 \loss}{\partial w_i\partial w_j\partial w_k}(x+c\Delta h)\enspace ,
\]
where $w\in\R^d$ with $d=km$ is the vectorization of $\w\in\R^{m\times k}$. 
We now bound this remainder. Using the representation of the loss Hessian from Eq.~\ref{eq:simplifiedHessian}, we can write the partial third derivatives in the remainder as
\[
\frac{\partial^3 \loss}{\partial w_i\partial w_j\partial w_k}(x)=\sum_{\stackrel{o,l,j\in [k]}{a,b,c\in [m]}}-\left(\hat{y}_j\hat{y}_o(\mathds{1}_{o=j}-\hat{y}_l)+\hat{y}_l\hat{y}_o(\mathds{1}_{o=l}-\hat{y}_j)\right)\phi(x)_a\phi(x)_b\phi(x)_c\enspace ,
\]
where $\hat{y}=f(x)$. Under the assumption that $\phi$ is $L$-Lipschitz, for all $x\in\X$, $\|x\|\leq 1$ and observing that $\sum_{o\in [k]}\hat{y}_o=1$ we can bound this term by
\begin{equation}
\frac{\partial^3 \loss}{\partial w_i\partial w_j\partial w_k}(x)\leq \frac{1}{4}kmL'^3\enspace .
\label{eq:boundThirdDerivative}
\end{equation}
The terms $k,m$ follow from the sum over all rows and columns of $\w$, and the factor $4^{-1}$ follows from the fact that the predictions in $\hat{y}$ sum up to $1$. The factor $L'^3$ can be derived as follows.  
\begin{align}
    \phi(x)_i \leq ||\phi(x)_i|| &= || (\phi(x)_i - \phi(\mathbf{0})_i) + \phi(\mathbf{0})_i|| &&\\
                                &  \leq  || (\phi(x)_i - \phi(\mathbf{0})_i)|| + ||\phi(\mathbf{0})_i|| &&\\
                                &  \leq  L||x-\mathbf{0} || + ||\phi(\mathbf{0})_i|| && (\phi \text{ is } L\text{-Lipschitz})\\
                                &  \leq  L + ||\phi(\mathbf{0})_i|| \leq L + C_{\phi(\mathbf{0})} && (||x||\leq1)\\
\end{align}
where $C_{\phi(\mathbf{0})}:= max_i  ||\phi(\mathbf{0})_i||$. For Relu networks without bias $C_{\phi(\mathbf{0})}$ is 0 and
empirically for networks with a bias term, it is very small i.e. $\approx1$. Therefore $L'=L+C_{\phi(\mathbf{0})}\approx L$, in particular since for most realistic neural networks $L$ is large. For simplicity, we subsitute $L=L'$.

Inserting this bound into Eq.~\ref{eq:lossBoundTaylor} yields 
\[
\left|\loss(f(\xi),y) - \loss(f(x),y)\right|\leq \Delta\|\w\|_F\|\nabla_\w\loss(\w)\|_F + \frac{\Delta^2}{2}\kappa^\phi_{Tr}(\w) + \frac{\Delta^3}{3!}\frac{1}{4}kmL^3\enspace .
\]
The result follows from setting $\Delta\leq L\delta r^{-1}$.
\end{proof}

\subsection{Proof of Proposition~\ref{prop:adversarialRobustness}}
\label{appdx:proofPropAdvRobustness}
\begin{pr}
For a dataset $S\subset\X\times\Y$ with $\|x\|\leq 1$ for all $x\in\X$, a model $f(x)=g(\w\phi(x))$ at a minimum $\w\in\R^{m\times k}$ wrt. $S$, with $\phi$ $L$-Lipschitz and $\|\phi(x)\|\geq r$, and a loss function $\loss(\w)=\loss(g(\w \phi(x)),y)$ of $f$ on $(x,y)$, $d$ being the $L2$-distance, and $\epsilon>0$, $f$ is $(\epsilon,\delta,S)$-robust against adversarial examples with
\begin{equation*}
    \begin{split}
        \delta&\geq\left(-\frac{8r^3k^3m^3L^9+27\epsilon}{27L^3\kappa_{Tr}^\phi(\w)}+\left(-\frac{2^7}{27}\frac{r^6k^6m^6L^{3}}{\kappa_{Tr}^\phi(\w)^6}+\frac{r^6\epsilon^2}{L^6\kappa_{Tr}^\phi(\w)^2}-\frac{2^4r^6\epsilon k^3m^3L^{\frac32}}{27\kappa_{Tr}^\phi(\w)^4}\right)^\frac{1}{2}\right)^\frac{1}{3} \\
        &+ \left(-\frac{8r^3k^3m^3L^9+27\epsilon}{27L^3\kappa_{Tr}^\phi(\w)}-\left(-\frac{2^7}{27}\frac{r^6k^6m^6L^{3}}{\kappa_{Tr}^\phi(\w)^6}+\frac{r^6\epsilon^2}{L^6\kappa_{Tr}^\phi(\w)^2}-\frac{2^4r^6\epsilon k^3m^3L^{\frac32}}{27\kappa_{Tr}^\phi(\w)^4}\right)^\frac{1}{2}\right)^\frac{1}{3} \\
        &+\frac{2rkmL^2}{72\kappa_{Tr}^\phi(\w)}\\
    \end{split}
\end{equation*}
where $\kappa_{Tr}(\w)$ is the relative flatness of $f$ wrt. $\w$. That is,
\[
\delta\propto \frac{\epsilon^{\frac13}}{(L^3\kappa_{Tr}(\w))^{\frac13}} + \frac{rkmL^2}{\kappa_{Tr}(\w)}\enspace .
\]
\label{prop:adversarialRobustness:formal}
\end{pr}
\begin{proof}
From Prop.~\ref{prop:lossBoundAdversarial} it follows that we achieve $(\epsilon,\Delta,S)$-robustness where 
\[
\epsilon = \frac{\Delta^2}{2}\kappa^\phi_{Tr}(\w) + \frac{\Delta^3}{24}kmL^3\enspace .
\]
First, we need to solve this cubic equation for $\Delta$. For that, we substitute $a=\frac{kmL^3}{24}$ and $b=\frac{\kappa_{Tr}^\phi(w)}{2}$ and get
\[
0=a\Delta^3 + b\Delta^2 - \epsilon=\Delta^3+\frac{a}{b}\Delta^2 - \frac{\epsilon}{b}=\Delta^3+\alpha\Delta^2 -\beta\enspace ,
\]
by subsituting $\alpha=\frac{a}{b}$ and $\beta=\frac{\epsilon}{b}$. We use the depressed cubic form $\Delta=t-\frac{\alpha}{3}$ and get
\begin{equation*}
    \begin{split}
        &0=\left(t- \frac{\alpha}{3}\right)^3 + \alpha\left(t - \frac{\alpha}{3}\right)^2 - \beta\\
        \Leftarrow&0=t^3-t^2\alpha+t\frac{\alpha^2}{3}-\frac{\alpha^3}{27}+t^2\alpha -\frac{2\alpha^2t}{3}+\frac{\alpha^3}{9}-\beta\\
        \Leftarrow&0=t^3-t\frac{\alpha^2}{3}+\frac{2\alpha^3}{27}-\beta\enspace .
    \end{split}
\end{equation*}
with $p=-\frac{\alpha^2}{3}$ and $q=\frac{2\alpha^3}{27}-\beta$ we get the form $0=t^3+pt+q$ for which we can apply Cardano's formula.
\[
t = \left(-\frac{q}{2}+\left(\frac{q^2}{4}+\frac{p^3}{9}\right)^\frac{1}{2}\right)^\frac{1}{3} + \left(-\frac{q}{2}-\left(\frac{q^2}{4}+\frac{p^3}{9}\right)^\frac{1}{2}\right)^\frac{1}{3}
\]
Resubstituting $p,q,\alpha,\beta$ yields
\[
\frac{q}{2}=\frac{a^3}{27b^3}-\frac{\epsilon}{2n}\enspace ,\enspace \frac{p^3}{9} = -\frac{1}{9^3}\frac{a^6}{b^6}\enspace ,\enspace \frac{q^2}{4}=\frac{1}{27^2}\frac{a^6}{b^6}+\frac{\epsilon^2}{4b^2}-\frac{\alpha^3\epsilon}{27b^4}\enspace .
\]
Substituting this in the solution for $t$ then gives
\begin{equation*}
    \begin{split}
t &= \left(-\frac{a^3}{27b^3}+\frac{\epsilon}{2b}+\left(-\frac{2a^6}{27b^6}+\frac{\epsilon^2}{4b^2}-\frac{a^3\epsilon}{27b^4}\right)^\frac{1}{2}\right)^\frac{1}{3} \\
&+ \left(-\frac{a^3}{27b^3}+\frac{\epsilon}{2b}-
\left(-\frac{2a^6}{27b^6}+\frac{\epsilon^2}{4b^2}-\frac{a^3\epsilon}{27b^4}\right)^\frac{1}{2}\right)^\frac{1}{3} 
    \end{split}
\end{equation*}
Substituting $a,b$ and $t=\Delta+\frac{\alpha}{3}$ yields
\begin{equation*}
    \begin{split}
\Delta &= \left(-\frac{8k^3m^3L^9+27\epsilon}{27\kappa_{Tr}^\phi(\w)}+\left(-\frac{2^7}{27}\frac{k^6m^6L^{18}}{\kappa_{Tr}^\phi(\w)^6}+\frac{\epsilon^2}{\kappa_{Tr}^\phi(\w)^2}-\frac{2^4\epsilon k^3m^3L^{9}}{27\kappa_{Tr}^\phi(\w)^4}\right)^\frac{1}{2}\right)^\frac{1}{3} \\
&+ \left(-\frac{8k^3m^3L^9+27\epsilon}{27\kappa_{Tr}^\phi(\w)}-\left(-\frac{2^7}{27}\frac{k^6m^6L^{18}}{\kappa_{Tr}^\phi(\w)^6}+\frac{\epsilon^2}{\kappa_{Tr}^\phi(\w)^2}-\frac{2^4\epsilon k^3m^3L^{9}}{27\kappa_{Tr}^\phi(\w)^4}\right)^\frac{1}{2}\right)^\frac{1}{3}\\
&+\frac{2kmL^3}{72\kappa_{Tr}^\phi(\w)}
    \end{split}
\end{equation*}
Finally, from Lemma~\ref{lm:relationFeatureInput} we have $\Delta\leq L\delta r^{-1}$, so that 
\begin{equation*}
    \begin{split}
\delta &\geq \frac{r}{L}\left(-\frac{8k^3m^3L^9+27\epsilon}{27\kappa_{Tr}^\phi(\w)}+\left(-\frac{2^7}{27}\frac{k^6m^6L^{18}}{\kappa_{Tr}^\phi(\w)^6}+\frac{\epsilon^2}{\kappa_{Tr}^\phi(\w)^2}-\frac{2^4\epsilon k^3m^3L^{9}}{27\kappa_{Tr}^\phi(\w)^4}\right)^\frac{1}{2}\right)^\frac{1}{3} \\
&+ \frac{r}{L}\left(-\frac{8k^3m^3L^9+27\epsilon}{27\kappa_{Tr}^\phi(\w)}-\left(-\frac{2^7}{27}\frac{k^6m^6L^{18}}{\kappa_{Tr}^\phi(\w)^6}+\frac{\epsilon^2}{\kappa_{Tr}^\phi(\w)^2}-\frac{2^4\epsilon k^3m^3L^{9}}{27\kappa_{Tr}^\phi(\w)^4}\right)^\frac{1}{2}\right)^\frac{1}{3}\\
&+\frac{2rkmL^2}{72\kappa_{Tr}^\phi(\w)}\\
=&\left(-\frac{8r^3k^3m^3L^9+27\epsilon}{27L^3\kappa_{Tr}^\phi(\w)}+\left(-\frac{2^7}{27}\frac{r^6k^6m^6L^{3}}{\kappa_{Tr}^\phi(\w)^6}+\frac{r^6\epsilon^2}{L^6\kappa_{Tr}^\phi(\w)^2}-\frac{2^4r^6\epsilon k^3m^3L^{\frac32}}{27\kappa_{Tr}^\phi(\w)^4}\right)^\frac{1}{2}\right)^\frac{1}{3} \\
&+ \left(-\frac{8r^3k^3m^3L^9+27\epsilon}{27L^3\kappa_{Tr}^\phi(\w)}-\left(-\frac{2^7}{27}\frac{r^6k^6m^6L^{3}}{\kappa_{Tr}^\phi(\w)^6}+\frac{r^6\epsilon^2}{L^6\kappa_{Tr}^\phi(\w)^2}-\frac{2^4r^6\epsilon k^3m^3L^{\frac32}}{27\kappa_{Tr}^\phi(\w)^4}\right)^\frac{1}{2}\right)^\frac{1}{3} \\
&+\frac{2rkmL^2}{72\kappa_{Tr}^\phi(\w)}\\
    \end{split}
\end{equation*}
\end{proof}

\subsection{Derivation of Hessian and Third Derivative}
\label{app:hessian}
Let $\phi \in \R^{m}$ denote the embedding of the feature extractor and $W\in \R^{K \times m}$, where we denote the weights of the 
$k$-th neuron as $w_k$. The output layer is given by the softmax function $\pred_k = \operatorname{softmax}(W\phi) \in \R^K$.
More precisely the softmax is given by,

$$ \pred_k = \frac{\exp(w_k \phi)}{\sum_{j=1}^K \exp(w_j \phi)}$$

For simplicity, we omit the bias term. The one-hot encoded ground truth is given by $y$. 
The derivative of the loss $L$ function wrt. the weight vector $w_j$ can be computed as

$$ \frac{\partial L(y,\hat{y})}{\partial w_j} = - (y_j - \pred_j)\phi^T$$

\paragraph{Second derivative}

\begin{align*}
    \frac{\partial L(y,\hat{y})}{\partial w_l \; \partial w_j} &= \frac{\partial}{\partial w_l}- (y_j - \pred_j)\phi^T \\
                                                               &= \partl{l} - y_j\phi^T  +  \partl{l} \pred_j     \phi^T  \\
                                                               &= \partl{l} \pred_j     \phi^T  
\end{align*} 
The last equation follows from $y$ being independent of $w_l$. Next, we do a case analysis on $l=j$.
\begin{itemize}
    \item ( $l=j$): In (1), we use the quotient rule and in (2) definition of softmax. \begin{align*} 
                        \partl{l} \pred_j   \phi^T  &=  \partl{j} \frac{\exp(w_j \phi)}{\sum_{k=1}^K \exp(w_k \phi)} \phi^T \\
                                                  &= \frac{(\exp(w_j \phi) \sum_{k=1}^K \exp(w_k \phi)) \phi \phi^T - \exp(w_j \phi) \exp(w_j \phi) \phi \phi^T }{(\sum_{k=1}^K \exp(w_k \phi))^2} \; \text{(1)}  \\
                                                  &= \left (\frac{\exp(w_j \phi) \sum_{k=1}^K \exp(w_k \phi) }{(\sum_{k=1}^K \exp(w_k \phi))^2} -\frac{\exp(w_j \phi)^2 }{(\sum_{k=1}^K \exp(w_k \phi))^2}  \right)\phi \phi^T \\
                                                  &= (\pred_j - \pred_j^2)\phi \phi^T \in \R^{m\times m} \; \text{(2)}
                    \end{align*} 
     \item ( $l\neq j$): Again quotient rule, but the left side vanishes.
     \begin{align*}
          \partl{l} \pred_j   \phi^T  = - \pred_l \pred_j \phi \phi^T \in \R^{m\times m}
     \end{align*}
\end{itemize}
Then we have
$$\frac{\partial L(y,\hat{y})}{\partial w_l \; \partial w_j} = \pred_l (\mathbbm{1}_{[l=j]}- \pred_j) \phi \phi^T \in \R^{m\times m}$$

The hessian is then given by
$$ H(L;W)(y,\pred)=(\text{diag}(\pred) - \pred \pred^T  ) \otimes \phi \phi^T \;\in \R^{Km\times Km}  $$

\paragraph{Third derivative}
First, rewrite 
\begin{align*}
    \frac{\partial L(y,\hat{y})}{\partial w_l \; \partial w_j} = \pred_l (\mathbbm{1}_{[l=j]}- \pred_j) \phi \phi^T =  \pred_l \mathbbm{1}_{[l=j]} \phi \phi^T - \pred_l \pred_j \phi \phi^T
\end{align*}
Then we define a new operator $ \pro: \R^n \times \R^m \times \R^o \rightarrow \R^{n\times m \times o}, \pro(x,y,z)_{ijk} = x_iy_j z_k$. 
We now compute 
 $$\frac{\partial L(y,\hat{y})}{\partial w_o\,\partial w_l \, \partial w_j} = \partl{o}  \pred_l \mathbbm{1}_{[l=j]} \phi \phi^T - \pred_l \pred_j \phi \phi^T$$

Again we make a CA on $j=l$
\begin{itemize}
    \item ( $l=j$): \begin{align*}
        \partl{o} ( \pred_l \phi \phi^T - \pred_l ^2 \phi \phi^T )&= \partl{o}  \pred_l \phi \phi^T - \partl{o}  \pred_l ^2 \phi \phi^T \\
                                                                  &= \pred_o (\mathbbm{1}_{[o=l]}- \pred_l) \pro(\phi,\phi,\phi) - 2   (\pred_o (\mathbbm{1}_{[o=l]}- \pred_l) \pro(\phi,\phi,\phi
                                                                  )) \\
                                                                  &=  - \pred_o (\mathbbm{1}_{[o=l]}- \pred_l) \pro(\phi,\phi,\phi)
    \end{align*}
    \item ( $l\neq j$):
    \begin{align*}
         \partl{o} - \pred_l \pred_j \phi \phi^T &= - (\partl{o} \pred_l) \pred_j \phi \phi^T -  \pred_l (\partl{o} \pred_j)\phi \phi^T \\
                                                 &=  - \pred_j \pred_o (\mathbbm{1}_{[o=l]}- \pred_l) \cdot \pro(\phi,\phi,\phi) -  \pred_l \pred_o(\mathbbm{1}_{[o=i]}- \pred_j) \cdot \pro(\phi,\phi,\phi) \\
                                                 &=  - [\pred_j \pred_o (\mathbbm{1}_{[o=l]}- \pred_l)  +  \pred_l \pred_o(\mathbbm{1}_{[o=i]}- \pred_j) ] \cdot \pro(\phi,\phi,\phi) \in \R^{m\times m\times m}
    \end{align*}
    $\rightarrow - [\pred_j \pred_o (\mathbbm{1}_{[o=l]}- \pred_l)  +  \pred_l \pred_o(\mathbbm{1}_{[o=j]}- \pred_j) ]_{j,l,o=1..k} \otimes \pro(\phi,\phi,\phi) \in \R^{Km\times Km\times Km}$
\end{itemize}
\section{Unnormalized Plots}
\label{app:plots}
\input{latex_figs/first-plot}
\input{latex_figs/llm-rfm}

\section{Code for experiments}
We use code from several resources, which we disclose here. First, the basis for training and attacking the CNNs stems from \citep{sehwag2020hydra}.
We modified the code according to our needs. The code for DenseNet121 stems from the official PyTorch library. To attack and 
evaluate the LLMs, we use the official implementation of the attack \citep{zou2023universal}. CIFAR-10 and CIFAR-100 were also 
downloaded from PyTorch.

\input{latex_figs/diff_layer}

\vfill
\section{Proof that Definition~\ref{def:advEx} Generalizes Definition~\ref{def:advEx-classic} }
\label{appdx:proof:defsAreEquivalent}
In the following, we provide a straightforward proof that Definition~\ref{def:advEx} is a generalization of the classical definition of adversarial examples by \citet{szegedy2014intriguing, papernot2016limitations} and \citet{carlini2017towards}. For that, we first restate our definition.
\begin{customdef}{2}
Let $\D$ be a distribution over an input space $\X$ and a label space $\Y$ with corresponding probability density function $P(X,Y)=P(Y\mid X)P(X)$.
Let $\loss:\Y\times\Y\rightarrow\R_+$ be a loss function, $f\in\F$ a model, and $(x,y)\in\X\times\Y$ be an example drawn according to $\D$. Given a distance function $d:\X\times\X\rightarrow\R_+$ over $\X$ and two thresholds $\epsilon,\delta\geq 0$, we call $\xi\in\X$ an \defemph{adversarial example} for $x$ if $d(x,\xi)\leq\delta$ and
\[
\expec{y_\xi\sim P(Y\mid X=\xi)}{\loss(f(\xi),y_\xi)} - \loss(f(x),y) > \epsilon \; .
\]
\end{customdef}
We now restate the classical definition here. For that, note that in ~\citet{szegedy2014intriguing}, the adversarial examples are assumed to be in $x+r\in [0,1]^m$ since they assume data to be images with pixel values in $[0,1]$. The original definition in \citet{szegedy2014intriguing} has an inconsistency, assuming $x\in\R^m$. For correctness, we therefore assume $x\in [0,1]^m$---Def.~\ref{def:advEx} would similarly generalize to arbitrary $\X$.
\begin{customdef}{1}[\citet{szegedy2014intriguing, papernot2016limitations} and \citet{carlini2017towards}(targeted)]
Let $f:\R^m\rightarrow\{1,\dots,k\}$ be a classifier, $x\in [0,1]^m$, and $l\in [k]$ with $l\neq f(x)$ a target class. Then for every
\[
r^* = \arg\min_{r\in\R^m}\|r\|_2\text{ s.t. }f(x+r)=l\text{ and }x+r\in [0,1]^m
\]
the perturbation $x+r^*$ is called an adversarial example. 
\end{customdef}
\begin{proof}

    We now prove that for particular choices of loss function, distance measure and thresholds, adversarial example fits to our general Definition~\ref{def:advEx} if and only if it fits to the classical Definition~\ref{def:advEx-classic}.

    For that, let $\X=[0,1]^m$, $\Y=\{1,\dots,k\}$ and for a distribution $\D$ and classifier $f$, we assume for $x\in\X$ that $P(Y=y|X=x)=1$, if $y=f(x)$ and $0$ otherwise. For $l\in [k]$, we set $\epsilon > 0$, and 
    \[
    \loss(\widehat{y},y)=\begin{cases}0\text{ , if }\widehat{y}\neq l\\\epsilon + 1\text{ , otherwise}\end{cases}\enspace 
    \]
    Furthermore, let $d=\|\cdot\|_2$ and
    \[
    \delta = \min_{r\in\R^m}\|r\|_2\text{ s.t. }\loss(f(x+r),f(x))>0\text{ and }x+r\in [0,1]^m\enspace 
    \]
    Lastly, we assume locally constant labels, i.e., for all $\xi=x+r$ with small perturbation $r$ it holds that $P(Y=y|X=x+r)=P(Y=y|X=x)$, i.e., the conditional distribution of the true label is constant around $x\in\X$.
    \begin{itemize}
        \item[(I)] Let $\xi=x+r^*$ be an adversarial example according to Def.~\ref{def:advEx}, then by construction of the loss function, $f(\xi)=l$. Furthermore, $\delta$ is chosen so that $\xi-x=r\in [0,1]^m$, and $\delta=\|r^*\|_2=\min_{r\in\R^m}\|r\|_2$ s.t. $f(x+r)=l$. Therefore, $x+r^*$ is also an adversarial example according to Def.~\ref{def:advEx-classic}.
        \item[(II)] Let $x+r^*$ be an adversarial example according to Def.~\ref{def:advEx-classic}, then $f(x+r^*)=l$ and thus $\loss(f(\xi),y_\xi)=\loss(f(x+r^*),y)=\loss(f(x+r^*),f(x))=\epsilon + 1 > \epsilon$. Furthermore $d(x,\xi)=\|x - x + r^*\|_2 = \min_{r\in\R^m}\|r\|_2\text{ s.t. }\loss(f(x+r),f(x))>0\text{ and }r\in [0,1]^m\leq \delta$.
    \end{itemize}
\end{proof}
Definition~\ref{def:advEx} naturally expands Definition~\ref{def:advEx-classic} by taking into account other distance metrics and formalizing the thresholds that are not specified in the original definition. A threshold on the loss difference generalizes from classification tasks, where targeted attacks can be modeled by a specific loss function. Therefore, our definition allows identifying adversarial example with respect to loss and not with respect to prediction itself, which is closer to the construction of adversarial examples with loss maximization. A distance threshold captures the requirement that adversarial perturbations are \emph{imperceptible}. Using the minimum distance is reasonable from an optimization perspective, but it also highlights a flaw in the original definition: If the minimal perturbation is so large that is indeed perceptible (or, even worse, would change, e.g., the picture of a cat to that of a dog) then it would still be considered an adversarial example under Def.~\ref{def:advEx-classic}. 

It seems more natural to set a fixed threshold $\delta$ on the distance that captures what it means for a perturbation to be imperceptible, as in the definition of untargeted attacks proposed by \citet{carlini2017towards}. From all adversarial examples defined that way, one can then find the closest to the clean example. We show that Definition~\ref{def:advEx} is also a generalization of this definition by \citet{carlini2017towards} of untargeted attacks with general $L_p$-distances.
\begin{df}[\citet{carlini2017towards} (untargeted)]
For $\X=\R^n$ and $\Y=\{1,\dots,m\}$, let $f:\X\rightarrow\Y$ be a classifier, $x\in\R^n$, $\|\cdot\|_p$ be a p-norm, and $\delta> 0$. Then every $\xi\in\X$ with $\|x-\xi\|_p\leq\delta$ is an adversarial example for $x$ if $f(\xi)\neq f(x)$.
\label{def:cw_untargeted}
\end{df}
\begin{proof}
    We assume $\X=\R^n$ and $\Y=\{1,\dots,m\}$, $d=\|\cdot\|_p$, and for a distribution $\D$ and classifier $f$, we assume for $x\in\X$ that $P(Y=y|X=x)=1$, if $y=f(x)$ and $0$ otherwise. For $\epsilon>0$, let $\loss$ be a loss function with $\loss(y,y)=0$ and for every $y'\neq y$, $\loss(y',y)>\epsilon$.  Lastly, we again assume a locally constant labels, i.e., for all $\xi\in\X$ with $\|x-\xi\|_p\leq\delta$ it holds that $P(Y=y|X=\xi)=P(Y=y|X=x)$. 
    \begin{itemize}
        \item[(I)] Let $\xi$ be an adversarial example according to Def.~\ref{def:advEx}, then by construction of the loss function, $f(\xi)\neq f(x)$. Furthermore, $\|x-\xi\|_p=d(x,\xi)\leq \delta$. Therefore, $\xi$ is also an adversarial example according to Def.~\ref{def:cw_untargeted}.
        \item[(II)] Let $\xi$ be an adversarial example according to Def.~\ref{def:cw_untargeted}. Then $d(x,\xi)=\|x-\xi\|_p\leq \delta$ and $\loss(f(\xi),y_\xi)-\loss(f(x),y)=\loss(f(\xi),f(x))-\loss(f(x),f(x))=\loss(f(\xi),f(x))>\epsilon$. Therefore, $\xi$ is also an adversarial example according to Def.~\ref{def:advEx}.
    \end{itemize}
\end{proof}

\section{Detecting Adversarial examples}
\label{app:detection}
It is possible to detect adversarial examples using a simple threshold on the relative sharpness measure. We did not include a practical study of this since it would go beyond the scope of this paper. Developing a sound method requires more than fine-tuning the threshold. Moreover, this requires comparing the approach to a wide range of state-of-the-art detection methods, which would be a paper of its own. Therefore, we leave this interesting practical aspect for future work. Nonetheless, we provide preliminary results. We  trained a decision stump on the sharpness of clean and adversarial samples on CIFAR-10 for \wrn using 5-fold cross-validation, which yields the following accuracies: $[0.92, 0.92, 0.93, 0.92, 0.92] $, i.e., adversarial examples can be detected with an average accuracy of $0.92$ with
little to no difference between the folds.

%% file: latex_figs/first-plot.tex
\begin{figure}[h!]
\centering
\begin{subfigure}[h]{0.4\linewidth}
        \begin{tikzpicture}
                \usetikzlibrary{calc}
                \begin{axis}[ 
                    smalljonas line,
                    cycle list name = prcl-line,
                    legend style={yshift=1.0cm,font=\scriptsize,xshift=4cm},
                    width=\linewidth,
                    height=3cm,
                    ylabel={Relative sharpness},
                    xlabel={Attack iteration},
                    pretty labelshift,
                    line width=0.75,
                    legend columns = 9,
                    tick label style={/pgf/number format/fixed},
                    legend entries = {\wrn,\resnet,\vgg,\dense}
                ]
                \foreach \x in {wrn, resnet,vgg11_bn,dense121}{
                    \addplot+[forget plot, eda errorbarcolored, y dir=plus, y explicit]
                    table[x={0}, y={trace-\x-cifar10}, y error expr=\thisrow{trace-\x-cifar10}-\thisrow{stdtrace-\x-cifar10}, col sep=comma]{expres/debug-rfm-for-clean.csv};
                    \addplot+[eda errorbarcolored, y dir=minus, y explicit]
                    table[x={0}, y={trace-\x-cifar10}, y error expr=\thisrow{trace-\x-cifar10}-\thisrow{stdtrace-\x-cifar10}, col sep=comma]{expres/debug-rfm-for-clean.csv};
                }    
                \end{axis}
            \end{tikzpicture}
                    \subcaption{Sharpness CIFAR-10}
\end{subfigure}%
\begin{subfigure}[h]{0.4\linewidth}
        \begin{tikzpicture}
                \usetikzlibrary{calc}
                \begin{axis}[ 
                    smalljonas line,
                    cycle list name = prcl-line,
                    legend style={yshift=1.0cm,font=\scriptsize,xshift=6cm},
                    width=\linewidth,
                    height=3cm,
                    ylabel={Loss},
                    xlabel={Attack iteration},
                    pretty labelshift,
                    line width=0.75,
                    legend columns = 9,
                    tick label style={/pgf/number format/fixed},
                ]
                \foreach \x in {wrn, resnet,vgg11_bn,dense121}{
                    \addplot+[forget plot, eda errorbarcolored, y dir=plus, y explicit]
                    table[x={0}, y={loss-\x-cifar10}, y error expr=\thisrow{loss-\x-cifar10}-\thisrow{stdloss-\x-cifar10}, col sep=comma]{expres/debug-rfm-for-clean.csv};
                    \addplot+[eda errorbarcolored, y dir=minus, y explicit]
                    table[x={0}, y={loss-\x-cifar10}, y error expr=\thisrow{loss-\x-cifar10}-\thisrow{stdloss-\x-cifar10}, col sep=comma]{expres/debug-rfm-for-clean.csv};%
                }    
                \end{axis}
            \end{tikzpicture}
                    \subcaption{Loss CIFAR-10}
\end{subfigure}
\begin{subfigure}[h]{0.4\linewidth}
        \begin{tikzpicture}
                \usetikzlibrary{calc}
                \begin{axis}[ 
                    smalljonas line,
                    cycle list name = prcl-line,
                    legend style={yshift=1.0cm,font=\scriptsize,xshift=6cm},
                    width=\linewidth,
                    height=3cm,
                    ylabel={Relative sharpness},
                    xlabel={Attack iteration},
                    pretty labelshift,
                    line width=0.75,
                    legend columns = 9,
                    tick label style={/pgf/number format/fixed},
                ]
                \foreach \x in {wrn, resnet,vgg11_bn,dense121}{
                    \addplot+[forget plot, eda errorbarcolored, y dir=plus, y explicit]
                    table[x={0}, y={trace-\x-cifar100}, y error expr=\thisrow{trace-\x-cifar100}-\thisrow{stdtrace-\x-cifar100}, col sep=comma]{expres/debug-rfm-for-clean.csv};
                    \addplot+[eda errorbarcolored, y dir=minus, y explicit]
                    table[x={0}, y={trace-\x-cifar100}, y error expr=\thisrow{trace-\x-cifar100}-\thisrow{stdtrace-\x-cifar100}, col sep=comma]{expres/debug-rfm-for-clean.csv};%
                }    
                \end{axis}
            \end{tikzpicture}
            \subcaption{Sharpness CIFAR-100}
\end{subfigure}%
\begin{subfigure}[h]{0.4\linewidth}
        \begin{tikzpicture}
                \usetikzlibrary{calc}
                \begin{axis}[ 
                    smalljonas line,
                    cycle list name = prcl-line,
                    legend style={yshift=1.0cm,font=\scriptsize,xshift=6cm},
                    width=\linewidth,
                    height=3cm,
                    ylabel={Loss},
                    xlabel={Attack iteration},
                    pretty labelshift,
                    line width=0.75,
                    legend columns = 9,
                    tick label style={/pgf/number format/fixed},
                ]
                \foreach \x in {wrn, resnet,vgg11_bn,dense121}{
                    \addplot+[forget plot, eda errorbarcolored, y dir=plus, y explicit]
                    table[x={0}, y={loss-\x-cifar100}, y error expr=\thisrow{loss-\x-cifar100}-\thisrow{stdloss-\x-cifar100}, col sep=comma]{expres/debug-rfm-for-clean.csv};
                    \addplot+[eda errorbarcolored, y dir=minus, y explicit]
                    table[x={0}, y={loss-\x-cifar100}, y error expr=\thisrow{loss-\x-cifar100}-\thisrow{stdloss-\x-cifar100}, col sep=comma]{expres/debug-rfm-for-clean.csv};%
                }    
                \end{axis}
            \end{tikzpicture}
            \subcaption{Loss CIFAR-100}
\end{subfigure}
\caption{We report the relative sharpness on the attack trajectory of attack for \wrn, \resnet, \vgg and \dense
on the test set of CIFAR-10 \& CIFAR-100. We observe that adversarial examples first reach a sharp region, but with strength of the attack increasing they are in very flat region. We also display the standard deviation of the values on individual inputs.}
\label{fig:first}
\end{figure}
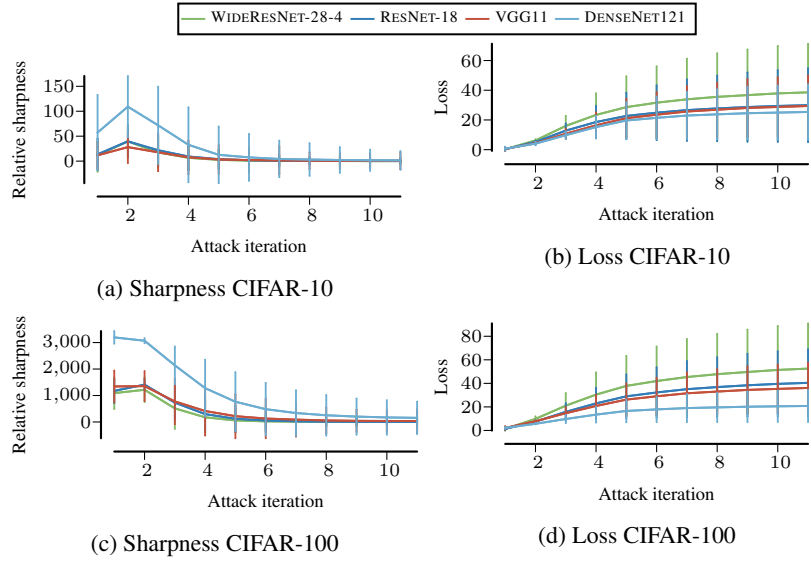

%% file: latex_figs/llm-rfm.tex
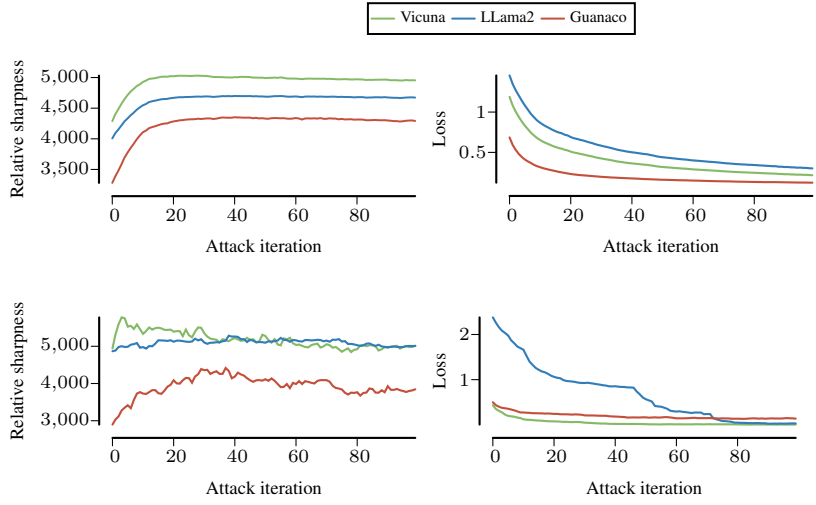
\begin{figure}
\centering
\begin{subfigure}[t]{0.4\linewidth}
        \begin{tikzpicture}
                \usetikzlibrary{calc}
                \begin{axis}[ 
                    smalljonas line,
                    cycle list name = prcl-line,
                    legend style={yshift=1.0cm,font=\scriptsize,xshift=3cm},
                    width=\linewidth,
                    height=3cm,
                    ylabel={Relative sharpness},
                    xlabel={Attack iteration},
                    pretty labelshift,
                    line width=0.75,
                    legend columns = 9,
                    tick label style={/pgf/number format/fixed},
                    legend entries = {Vicuna, LLama2, Guanaco},
                ]
                \foreach \x in {vicuna, llama2, guanaco}{
                    \addplot table[x={index},y={\x-fam},col sep=comma]{expres/llms/\x-attack.csv};%
                }    
                \end{axis}
            \end{tikzpicture}
                \label{fig:llm-adv-a}
\end{subfigure}%
\begin{subfigure}[t]{0.4\linewidth}
        \begin{tikzpicture}
                \usetikzlibrary{calc}
                \begin{axis}[ 
                    smalljonas line,
                    cycle list name = prcl-line,
                    legend style={yshift=0.7cm,font=\scriptsize,xshift=3.4cm},
                    width=\linewidth,
                    height=3cm,
                    ylabel={Loss},
                    xlabel={Attack iteration},
                    pretty labelshift,
                    legend columns = 5,
                    tick label style={/pgf/number format/fixed}
                ]
                \foreach \x in {vicuna, llama2, guanaco}{
                    \addplot table[x={index},y={\x-loss},col sep=comma]{expres/llms/\x-attack.csv};%
                }       
                \end{axis}
            \end{tikzpicture}
                \label{fig:llm-adv-b}
\end{subfigure}
\begin{subfigure}[t]{0.4\linewidth}
        \begin{tikzpicture}
                \usetikzlibrary{calc}
                \begin{axis}[ 
                    smalljonas line,
                    cycle list name = prcl-line,
                    legend style={yshift=0.7cm,font=\scriptsize,xshift=3.4cm},
                    width=\linewidth,
                    height=3cm,
                    ylabel={Relative sharpness},
                    xlabel={Attack iteration},
                    pretty labelshift,
                    legend columns = 5,
                    tick label style={/pgf/number format/fixed}
                ]
                \foreach \x in {vicuna, llama2, guanaco}{
                    \addplot table[x={index},y={fam_pos},col sep=comma]{expres/posneg/\x_posneg.csv};%
                }       
                \end{axis}
            \end{tikzpicture}
            \label{fig:llm-adv-c}
\end{subfigure}%
\begin{subfigure}[t]{0.4\linewidth}
        \begin{tikzpicture}
                \usetikzlibrary{calc}
                \begin{axis}[ 
                    smalljonas line,
                    cycle list name = prcl-line,
                    legend style={yshift=0.7cm,font=\scriptsize,xshift=3.4cm},
                    width=\linewidth,
                    height=3cm,
                    ylabel={Loss},
                    xlabel={Attack iteration},
                    pretty labelshift,
                    legend columns = 5,
                    tick label style={/pgf/number format/fixed}
                ]
                \foreach \x in {vicuna, llama2, guanaco}{
                    \addplot table[x={index},y={loss_pos},col sep=comma]{expres/posneg/\x_posneg.csv};%
                }       
                \end{axis}
            \end{tikzpicture}
            \label{fig:llm-adv-d}
\end{subfigure}
\caption{\textbf{Top}. We plot the relative sharpness and loss of the adversarial prompt for Viucna, LLama2 and Guanaco when attacked by
the method of \cite{zou2023universal}. \textbf{Bottom}.We give per model example trajectories that  first get sharper and then flatter again.}
\label{fig:llm-adv}
\end{figure}

%% file: latex_figs/diff_layer.tex
\begin{figure}[ht]
\begin{subfigure}[t]{0.2\linewidth}
        \begin{tikzpicture}
                \usetikzlibrary{calc}
                \begin{axis}[ 
                    smalljonas line,
                    cycle list name = more-prcl,
                    legend style={yshift=1cm,font=\scriptsize,xshift=9cm},
                    width=1.08\linewidth,
                    height=2.9cm,
                    ylabel={Relative flatness},
                    xlabel={Attack iteration},
                    pretty labelshift,
                    line width=0.75,
                    ymin=0,
                    ymax=50,
                    xmin=0,
                    xmax=10,
                    legend columns = 9,
                    tick label style={/pgf/number format/fixed},
                ]
                    \addplot table[x={0},y={layer_103},col sep=comma]{expres/sharpness_per_layer.csv};%

                \end{axis}
            \end{tikzpicture}
            \subcaption{Layer $l$}
\end{subfigure}%
\begin{subfigure}[t]{0.2\linewidth}
        \begin{tikzpicture}
                \usetikzlibrary{calc}
                \begin{axis}[ 
                    smalljonas line,
                    cycle list name = more-prcl,
                    legend style={yshift=1cm,font=\scriptsize,xshift=9cm},
                    width=1.08\linewidth,
                    height=2.9cm,
                    ylabel={Relative flatness},
                    xlabel={Attack iteration},
                    pretty labelshift,
                    line width=0.75,
                    ymin=0,
                    ymax=240,
                    xmin=0,
                    xmax=10,
                    legend columns = 9,
                    tick label style={/pgf/number format/fixed},
                ]
                    \addplot table[x={0},y={layer_100},col sep=comma]{expres/sharpness_per_layer.csv};%

                \end{axis}
            \end{tikzpicture}
            \subcaption{Layer $l-1$}
\end{subfigure}%
\begin{subfigure}[t]{0.2\linewidth}
        \begin{tikzpicture}
                \usetikzlibrary{calc}
                \begin{axis}[ 
                    smalljonas line,
                    cycle list name = more-prcl,
                    legend style={yshift=1cm,font=\scriptsize,xshift=9cm},
                    width=1.08\linewidth,
                    height=2.9cm,
                    ylabel={Relative flatness},
                    xlabel={Attack iteration},
                    pretty labelshift,
                    line width=0.75,
                    ymin=0,
                    ymax=800,
                    xmin=0,
                    xmax=10,
                    legend columns = 9,
                    tick label style={/pgf/number format/fixed},
                ]
                    \addplot table[x={0},y={layer_97},col sep=comma]{expres/sharpness_per_layer.csv};%

                \end{axis}
            \end{tikzpicture}
                        \subcaption{Layer $l-2$}
                        \label{fig:adv-train-c-layers}
\end{subfigure}%
\begin{subfigure}[t]{0.2\linewidth}
        \begin{tikzpicture}
                \usetikzlibrary{calc}
                \begin{axis}[ 
                    smalljonas line,
                    cycle list name = more-prcl,
                    legend style={yshift=1cm,font=\scriptsize,xshift=9cm},
                    width=1.08\linewidth,
                    height=2.9cm,
                    ylabel={Relative flatness},
                    xlabel={Attack iteration},
                    pretty labelshift,
                    line width=0.75,
                    ymin=0,
                    ymax=700,
                    xmin=0,
                    xmax=10,
                    legend columns = 9,
                    tick label style={/pgf/number format/fixed},
                ]
                    \addplot table[x={0},y={layer_94},col sep=comma]{expres/sharpness_per_layer.csv};%

                \end{axis}
            \end{tikzpicture}
                        \subcaption{Layer $l-3$}
\end{subfigure}%
\begin{subfigure}[t]{0.2\linewidth}
        \begin{tikzpicture}
                \usetikzlibrary{calc}
                \begin{axis}[ 
                    smalljonas line,
                    cycle list name = more-prcl,
                    legend style={yshift=1cm,font=\scriptsize,xshift=9cm},
                    width=1.08\linewidth,
                    height=2.9cm,
                    ylabel={Relative flatness},
                    xlabel={Attack iteration},
                    pretty labelshift,
                    line width=0.75,
                    ymin=0,
                    ymax=1500,
                    xmin=0,
                    xmax=10,
                    legend columns = 9,
                    yticklabels = {0,0,500, 1k, 1.5k},
                    tick label style={/pgf/number format/fixed},
                ]
                    \addplot table[x={0},y={layer_91},col sep=comma]{expres/sharpness_per_layer.csv};%

                \end{axis}
            \end{tikzpicture}
                        \subcaption{Layer $l-4$}
\end{subfigure}
\caption{We show the relative sharpness measure computed in the penultimate layer $l$ and in shallower layers $l-1$ to $l-4$ for
\wrn. Due to memory and runtime constraints, we
approximate the measure using Hutchinson trace estimation used in \cite{petzka2021relative} on 500 images. We observe
the same phenomena as in the penultimate layer, which justifies that we focus only on the penultimate layer for our theoretical
and experimental analysis.}
\label{fig:diff-layer}
\end{figure}
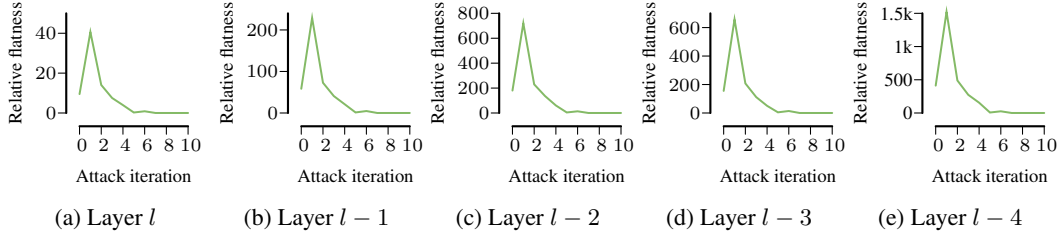

%% file: paper.bbl
\providecommand{\noopsort}[1]{}
\begin{thebibliography}{69}
\providecommand{\natexlab}[1]{#1}
\providecommand{\url}[1]{\texttt{#1}}
\expandafter\ifx\csname urlstyle\endcsname\relax
  \providecommand{\doi}[1]{doi: #1}\else
  \providecommand{\doi}{doi: \begingroup \urlstyle{rm}\Url}\fi

\bibitem[Adilova et~al.(2023)Adilova, Andriushchenko, Kamp, Fischer, and
  Jaggi]{adilova2023layer}
Linara Adilova, Maksym Andriushchenko, Michael Kamp, Asja Fischer, and Martin
  Jaggi.
\newblock Layer-wise linear mode connectivity.
\newblock In \emph{The Twelfth International Conference on Learning
  Representations}, 2023.

\bibitem[Alaifari et~al.(2018)Alaifari, Alberti, and
  Gauksson]{alaifari2018adef}
Rima Alaifari, Giovanni~S Alberti, and Tandri Gauksson.
\newblock Adef: an iterative algorithm to construct adversarial deformations.
\newblock In \emph{International Conference on Learning Representations}, 2018.

\bibitem[Andriushchenko et~al.(2020)Andriushchenko, Croce, Flammarion, and
  Hein]{andriushchenko2020square}
Maksym Andriushchenko, Francesco Croce, Nicolas Flammarion, and Matthias Hein.
\newblock Square attack: A query-efficient black-box adversarial attack via
  random search.
\newblock In \emph{European Conference on Computer Vision}, pp.\  484--501,
  2020.

\bibitem[Araujo et~al.(2023)Araujo, Havens, Delattre, Allauzen, and
  Hu]{araujounified}
Alexandre Araujo, Aaron~J Havens, Blaise Delattre, Alexandre Allauzen, and Bin
  Hu.
\newblock A unified algebraic perspective on lipschitz neural networks.
\newblock In \emph{The Eleventh International Conference on Learning
  Representations}, 2023.

\bibitem[Bakiskan et~al.(2022)Bakiskan, Cekic, and Madhow]{bakiskan2022early}
Can Bakiskan, Metehan Cekic, and Upamanyu Madhow.
\newblock Early layers are more important for adversarial robustness.
\newblock In \emph{ICLR 2022 Workshop on New Frontiers in Adversarial Machine
  Learning}, 2022.

\bibitem[Brown et~al.(2018)Brown, Mané, Roy, Abadi, and
  Gilmer]{brown_adversarial_2018}
Tom~B. Brown, Dandelion Mané, Aurko Roy, Martín Abadi, and Justin Gilmer.
\newblock Adversarial {Patch}, May 2018.
\newblock NIPS 2017, Long Beach, CA, USA.

\bibitem[Cai et~al.(2018)Cai, Liu, and Song]{cai2018curriculum}
Qi-Zhi Cai, Chang Liu, and Dawn Song.
\newblock Curriculum adversarial training.
\newblock In \emph{Proceedings of the 27th International Joint Conference on
  Artificial Intelligence}, pp.\  3740--3747, 2018.

\bibitem[Carlini \& Wagner(2017{\natexlab{a}})Carlini and
  Wagner]{carlini2017adversarial}
Nicholas Carlini and David Wagner.
\newblock Adversarial examples are not easily detected: Bypassing ten detection
  methods.
\newblock In \emph{Proceedings of the 10th ACM Workshop on Artificial
  Intelligence and Security}, pp.\  3--14, 2017{\natexlab{a}}.

\bibitem[Carlini \& Wagner(2017{\natexlab{b}})Carlini and
  Wagner]{carlini2017towards}
Nicholas Carlini and David Wagner.
\newblock Towards evaluating the robustness of neural networks.
\newblock In \emph{2017 ieee symposium on security and privacy (sp)}, pp.\
  39--57. IEEE, 2017{\natexlab{b}}.

\bibitem[Carmon et~al.(2019)Carmon, Raghunathan, Schmidt, Duchi, and
  Liang]{carmon2019unlabeled}
Yair Carmon, Aditi Raghunathan, Ludwig Schmidt, John~C Duchi, and Percy~S
  Liang.
\newblock Unlabeled data improves adversarial robustness.
\newblock \emph{Advances in neural information processing systems}, 32, 2019.

\bibitem[Chen et~al.(2023)Chen, Gao, Zhao, Ye, and Xu]{chen2023advdiffuser}
Xinquan Chen, Xitong Gao, Juanjuan Zhao, Kejiang Ye, and Cheng-Zhong Xu.
\newblock Advdiffuser: Natural adversarial example synthesis with diffusion
  models.
\newblock In \emph{Proceedings of the IEEE/CVF International Conference on
  Computer Vision}, pp.\  4562--4572, 2023.

\bibitem[Chiang et~al.(2023)Chiang, Li, Lin, Sheng, Wu, Zhang, Zheng, Zhuang,
  Zhuang, Gonzalez, Stoica, and Xing]{vicuna2023}
Wei-Lin Chiang, Zhuohan Li, Zi~Lin, Ying Sheng, Zhanghao Wu, Hao Zhang, Lianmin
  Zheng, Siyuan Zhuang, Yonghao Zhuang, Joseph~E. Gonzalez, Ion Stoica, and
  Eric~P. Xing.
\newblock Vicuna: An open-source chatbot impressing gpt-4 with 90\%* chatgpt
  quality, March 2023.

\bibitem[Cohen et~al.(2019)Cohen, Rosenfeld, and Kolter]{cohen2019certified}
Jeremy Cohen, Elan Rosenfeld, and Zico Kolter.
\newblock Certified adversarial robustness via randomized smoothing.
\newblock In \emph{International Conference on Machine Learning}, pp.\
  1310--1320, 2019.

\bibitem[Croce \& Hein(2020)Croce and Hein]{croce2020reliable}
Francesco Croce and Matthias Hein.
\newblock Reliable evaluation of adversarial robustness with an ensemble of
  diverse parameter-free attacks.
\newblock In \emph{International conference on machine learning}, pp.\
  2206--2216. PMLR, 2020.

\bibitem[Croce \& Hein(2021)Croce and Hein]{croce2021mind}
Francesco Croce and Matthias Hein.
\newblock Mind the box: $ l\_1 $-apgd for sparse adversarial attacks on image
  classifiers.
\newblock In \emph{International Conference on Machine Learning}, pp.\
  2201--2211. PMLR, 2021.

\bibitem[Dettmers et~al.(2024)Dettmers, Pagnoni, Holtzman, and
  Zettlemoyer]{dettmers2024qlora}
Tim Dettmers, Artidoro Pagnoni, Ari Holtzman, and Luke Zettlemoyer.
\newblock Qlora: Efficient finetuning of quantized llms.
\newblock \emph{Advances in Neural Information Processing Systems}, 36, 2024.

\bibitem[Dinh et~al.(2017)Dinh, Pascanu, Bengio, and Bengio]{dinh2017sharp}
Laurent Dinh, Razvan Pascanu, Samy Bengio, and Yoshua Bengio.
\newblock Sharp minima can generalize for deep nets.
\newblock In \emph{International Conference on Machine Learning}, pp.\
  1019--1028. PMLR, 2017.

\bibitem[Foret et~al.(2020)Foret, Kleiner, Mobahi, and
  Neyshabur]{foret2020sharpness}
Pierre Foret, Ariel Kleiner, Hossein Mobahi, and Behnam Neyshabur.
\newblock Sharpness-aware minimization for efficiently improving
  generalization.
\newblock In \emph{International Conference on Learning Representations}, 2020.

\bibitem[Goodfellow et~al.(2015)Goodfellow, Shlens, and
  Szegedy]{goodfellow2014explaining}
Ian~J Goodfellow, Jonathon Shlens, and Christian Szegedy.
\newblock Explaining and harnessing adversarial examples.
\newblock \emph{ICLR}, 2015.

\bibitem[Gouk et~al.(2021)Gouk, Frank, Pfahringer, and
  Cree]{gouk2021regularisation}
Henry Gouk, Eibe Frank, Bernhard Pfahringer, and Michael~J Cree.
\newblock Regularisation of neural networks by enforcing lipschitz continuity.
\newblock \emph{Machine Learning}, 110:\penalty0 393--416, 2021.

\bibitem[Gubri et~al.(2022)Gubri, Cordy, Papadakis, Traon, and
  Sen]{gubri2022lgv}
Martin Gubri, Maxime Cordy, Mike Papadakis, Yves~Le Traon, and Koushik Sen.
\newblock Lgv: Boosting adversarial example transferability from large
  geometric vicinity.
\newblock In \emph{European Conference on Computer Vision}, pp.\  603--618.
  Springer, 2022.

\bibitem[He et~al.(2016)He, Zhang, Ren, and Sun]{he2016deep}
Kaiming He, Xiangyu Zhang, Shaoqing Ren, and Jian Sun.
\newblock Deep residual learning for image recognition.
\newblock In \emph{Proceedings of the IEEE conference on computer vision and
  pattern recognition}, pp.\  770--778, 2016.

\bibitem[Hein \& Andriushchenko(2017)Hein and Andriushchenko]{hein2017formal}
Matthias Hein and Maksym Andriushchenko.
\newblock Formal guarantees on the robustness of a classifier against
  adversarial manipulation.
\newblock In \emph{Advances in Neural Information Processing Systems}, pp.\
  2266--2276, 2017.

\bibitem[Hochreiter \& Schmidhuber(1994)Hochreiter and
  Schmidhuber]{hochreiter1994simplifying}
Sepp Hochreiter and J{\"u}rgen Schmidhuber.
\newblock Simplifying neural nets by discovering flat minima.
\newblock \emph{Advances in neural information processing systems}, 7, 1994.

\bibitem[Hu et~al.(2019)Hu, Yu, Guo, Chao, and Weinberger]{hu2019new}
Shengyuan Hu, Tao Yu, Chuan Guo, Wei-Lun Chao, and Kilian~Q Weinberger.
\newblock A new defense against adversarial images: Turning a weakness into a
  strength.
\newblock \emph{Advances in neural information processing systems}, 32, 2019.

\bibitem[Huang et~al.(2017)Huang, Liu, Van Der~Maaten, and
  Weinberger]{huang2017densely}
Gao Huang, Zhuang Liu, Laurens Van Der~Maaten, and Kilian~Q Weinberger.
\newblock Densely connected convolutional networks.
\newblock In \emph{Proceedings of the IEEE conference on computer vision and
  pattern recognition}, pp.\  4700--4708, 2017.

\bibitem[Jiang et~al.(2019)Jiang, Neyshabur, Mobahi, Krishnan, and
  Bengio]{jiang2019fantastic}
Yiding Jiang, Behnam Neyshabur, Hossein Mobahi, Dilip Krishnan, and Samy
  Bengio.
\newblock Fantastic generalization measures and where to find them.
\newblock In \emph{International Conference on Learning Representations}, 2019.

\bibitem[Kanai et~al.(2023)Kanai, Yamada, Takahashi, Yamanaka, and
  Ida]{kanai2023relationship}
Sekitoshi Kanai, Masanori Yamada, Hiroshi Takahashi, Yuki Yamanaka, and
  Yasutoshi Ida.
\newblock Relationship between nonsmoothness in adversarial training,
  constraints of attacks, and flatness in the input space.
\newblock \emph{IEEE Transactions on Neural Networks and Learning Systems},
  2023.

\bibitem[Keskar et~al.(2016)Keskar, Mudigere, Nocedal, Smelyanskiy, and
  Tang]{keskar2016large}
Nitish~Shirish Keskar, Dheevatsa Mudigere, Jorge Nocedal, Mikhail Smelyanskiy,
  and Ping Tak~Peter Tang.
\newblock On large-batch training for deep learning: Generalization gap and
  sharp minima.
\newblock In \emph{International Conference on Learning Representations}, 2016.

\bibitem[Kumari et~al.(2019)Kumari, Singh, Sinha, Machiraju, Krishnamurthy, and
  Balasubramanian]{kumari2019harnessing}
Nupur Kumari, Mayank Singh, Abhishek Sinha, Harshitha Machiraju, Balaji
  Krishnamurthy, and Vineeth~N Balasubramanian.
\newblock Harnessing the vulnerability of latent layers in adversarially
  trained models.
\newblock In \emph{Proceedings of the 28th International Joint Conference on
  Artificial Intelligence}, pp.\  2779--2785, 2019.

\bibitem[Kurakin et~al.(2017)Kurakin, Goodfellow, and
  Bengio]{kurakin2017adversarial}
Alexey Kurakin, Ian~J Goodfellow, and Samy Bengio.
\newblock Adversarial machine learning at scale.
\newblock In \emph{Proceedings of the International Conference on Learning
  Representations}, 2017.

\bibitem[Lee et~al.(2018)Lee, Lee, Lee, and Shin]{lee2018simple}
Kimin Lee, Kibok Lee, Honglak Lee, and Jinwoo Shin.
\newblock A simple unified framework for detecting out-of-distribution samples
  and adversarial attacks.
\newblock \emph{Advances in neural information processing systems}, 31, 2018.

\bibitem[Li et~al.(2020)Li, Ma, Guo, Xue, and Qiu]{li2020bert}
Linyang Li, Ruotian Ma, Qipeng Guo, Xiangyang Xue, and Xipeng Qiu.
\newblock Bert-attack: Adversarial attack against bert using bert.
\newblock In \emph{Proceedings of the 2020 Conference on Empirical Methods in
  Natural Language Processing (EMNLP)}, pp.\  6193--6202, 2020.

\bibitem[Liang et~al.(2019)Liang, Poggio, Rakhlin, and Stokes]{liang2019fisher}
Tengyuan Liang, Tomaso Poggio, Alexander Rakhlin, and James Stokes.
\newblock Fisher-rao metric, geometry, and complexity of neural networks.
\newblock In \emph{The 22nd international conference on artificial intelligence
  and statistics}, pp.\  888--896. PMLR, 2019.

\bibitem[Liang \& Huang(2021)Liang and Huang]{liang2021large}
Youwei Liang and Dong Huang.
\newblock Large norms of cnn layers do not hurt adversarial robustness.
\newblock In \emph{Proceedings of the AAAI Conference on Artificial
  Intelligence}, volume~35, pp.\  8565--8573, 2021.

\bibitem[Ma et~al.(2018)Ma, Li, Wang, Erfani, Wijewickrema, Schoenebeck, Song,
  Houle, and Bailey]{ma2018characterizing}
Xingjun Ma, Bo~Li, Yisen Wang, Sarah~M Erfani, Sudanthi Wijewickrema, Grant
  Schoenebeck, Dawn Song, Michael~E Houle, and James Bailey.
\newblock Characterizing adversarial subspaces using local intrinsic
  dimensionality.
\newblock In \emph{International Conference on Learning Representations}, 2018.

\bibitem[Madry et~al.(2017)Madry, Makelov, Schmidt, Tsipras, and
  Vladu]{madry2017towards}
Aleksander Madry, Aleksandar Makelov, Ludwig Schmidt, Dimitris Tsipras, and
  Adrian Vladu.
\newblock Towards deep learning models resistant to adversarial attacks.
\newblock \emph{arXiv preprint arXiv:1706.06083}, 2017.

\bibitem[Mao et~al.(2018)Mao, Chen, Li, He, and Xue]{mao2019learning}
Xiaofeng Mao, Yuefeng Chen, Yuhong Li, Yuan He, and Hui Xue.
\newblock Learning to characterize adversarial subspaces.
\newblock In \emph{International Conference on Learning Representations}, 2018.

\bibitem[Moosavi-Dezfooli et~al.(2017)Moosavi-Dezfooli, Fawzi, Fawzi, and
  Frossard]{moosavi2017universal}
Seyed-Mohsen Moosavi-Dezfooli, Alhussein Fawzi, Omar Fawzi, and Pascal
  Frossard.
\newblock Universal adversarial perturbations.
\newblock In \emph{Proceedings of the IEEE conference on computer vision and
  pattern recognition}, pp.\  1765--1773, 2017.

\bibitem[Moosavi-Dezfooli et~al.(2019)Moosavi-Dezfooli, Fawzi, Uesato, and
  Frossard]{moosavi2019robustness}
Seyed-Mohsen Moosavi-Dezfooli, Alhussein Fawzi, Jonathan Uesato, and Pascal
  Frossard.
\newblock Robustness via curvature regularization, and vice versa.
\newblock In \emph{Proceedings of the IEEE Conference on Computer Vision and
  Pattern Recognition}, pp.\  9078--9086, 2019.

\bibitem[Narodytska \& Kasiviswanathan(2017)Narodytska and
  Kasiviswanathan]{narodytska_simple_2017}
Nina Narodytska and Shiva Kasiviswanathan.
\newblock Simple {Black}-{Box} {Adversarial} {Attacks} on {Deep} {Neural}
  {Networks}.
\newblock In \emph{2017 {IEEE} {Conference} on {Computer} {Vision} and
  {Pattern} {Recognition} {Workshops} ({CVPRW})}, pp.\  1310--1318, Honolulu,
  HI, USA, July 2017. IEEE.
\newblock ISBN 978-1-5386-0733-6.
\newblock \doi{10.1109/CVPRW.2017.172}.

\bibitem[Papernot et~al.(2016)Papernot, McDaniel, Jha, Fredrikson, Celik, and
  Swami]{papernot2016limitations}
Nicolas Papernot, Patrick McDaniel, Somesh Jha, Matt Fredrikson, Z~Berkay
  Celik, and Ananthram Swami.
\newblock The limitations of deep learning in adversarial settings.
\newblock In \emph{2016 IEEE European symposium on security and privacy
  (EuroS\&P)}, pp.\  372--387. IEEE, 2016.

\bibitem[Perolat et~al.(2018)Perolat, Malinowski, Piot, and
  Pietquin]{perolat_playing_2018}
Julien Perolat, Mateusz Malinowski, Bilal Piot, and Olivier Pietquin.
\newblock Playing the {Game} of {Universal} {Adversarial} {Perturbations},
  September 2018.
\newblock arXiv:1809.07802 [cs, stat].

\bibitem[Petzka et~al.(2021)Petzka, Kamp, Adilova, Sminchisescu, and
  Boley]{petzka2021relative}
Henning Petzka, Michael Kamp, Linara Adilova, Cristian Sminchisescu, and Mario
  Boley.
\newblock Relative flatness and generalization.
\newblock \emph{Advances in neural information processing systems},
  34:\penalty0 18420--18432, 2021.

\bibitem[Rahnama et~al.(2020)Rahnama, Nguyen, and Raff]{rahnama2020robust}
Arash Rahnama, Andre~T Nguyen, and Edward Raff.
\newblock Robust design of deep neural networks against adversarial attacks
  based on lyapunov theory.
\newblock In \emph{Proceedings of the IEEE/CVF Conference on Computer Vision
  and Pattern Recognition}, pp.\  8178--8187, 2020.

\bibitem[Salman et~al.(2019)Salman, Li, Razenshteyn, Zhang, Zhang, Bubeck, and
  Yang]{salman2019provably}
Hadi Salman, Jerry Li, Ilya Razenshteyn, Pengchuan Zhang, Huan Zhang, Sebastien
  Bubeck, and Greg Yang.
\newblock Provably robust deep learning via adversarially trained smoothed
  classifiers.
\newblock \emph{Advances in neural information processing systems}, 32, 2019.

\bibitem[Schmidt et~al.(2018)Schmidt, Santurkar, Tsipras, Talwar, and
  Madry]{schmidt2018adversarially}
Ludwig Schmidt, Shibani Santurkar, Dimitris Tsipras, Kunal Talwar, and
  Aleksander Madry.
\newblock Adversarially robust generalization requires more data.
\newblock In \emph{Advances in Neural Information Processing Systems}, pp.\
  5014--5026, 2018.

\bibitem[Sehwag et~al.(2020)Sehwag, Wang, Mittal, and Jana]{sehwag2020hydra}
Vikash Sehwag, Shiqi Wang, Prateek Mittal, and Suman Jana.
\newblock Hydra: Pruning adversarially robust neural networks.
\newblock \emph{Advances in Neural Information Processing Systems},
  33:\penalty0 19655--19666, 2020.

\bibitem[Shafahi et~al.(2019)Shafahi, Najibi, Ghiasi, Xu, Dickerson, Studer,
  Davis, Taylor, and Goldstein]{shafahi2019adversarial}
Ali Shafahi, Mahyar Najibi, Mohammad~Amin Ghiasi, Zheng Xu, John Dickerson,
  Christoph Studer, Larry~S Davis, Gavin Taylor, and Tom Goldstein.
\newblock Adversarial training for free!
\newblock \emph{Advances in neural information processing systems}, 32, 2019.

\bibitem[Shafahi et~al.(2020)Shafahi, Najibi, Xu, Dickerson, Davis, and
  Goldstein]{shafahi2020universal}
Ali Shafahi, Mahyar Najibi, Zheng Xu, John Dickerson, Larry~S Davis, and Tom
  Goldstein.
\newblock Universal adversarial training.
\newblock In \emph{Proceedings of the AAAI Conference on Artificial
  Intelligence}, volume~34, pp.\  5636--5643, 2020.

\bibitem[Simonyan \& Zisserman(2014)Simonyan and Zisserman]{Simonyan2014VeryDC}
Karen Simonyan and Andrew Zisserman.
\newblock Very deep convolutional networks for large-scale image recognition.
\newblock \emph{Internattional Conference on Machine Learning}, abs/1409.1556,
  2014.

\bibitem[Stutz et~al.(2021)Stutz, Hein, and Schiele]{stutz_relating_2021}
David Stutz, Matthias Hein, and Bernt Schiele.
\newblock Relating {Adversarially} {Robust} {Generalization} to {Flat}
  {Minima}.
\newblock In \emph{2021 {IEEE}/{CVF} {International} {Conference} on {Computer}
  {Vision} ({ICCV})}, pp.\  7787--7797, Montreal, QC, Canada, October 2021.
  IEEE.

\bibitem[Szegedy et~al.(2014)Szegedy, Zaremba, Sutskever, Bruna, Erhan,
  Goodfellow, and Fergus]{szegedy2014intriguing}
Christian Szegedy, Wojciech Zaremba, Ilya Sutskever, Joan Bruna, Dumitru Erhan,
  Ian Goodfellow, and Rob Fergus.
\newblock Intriguing properties of neural networks.
\newblock In \emph{International Conference on Learning Representations}, 2014.

\bibitem[Tian et~al.(2018)Tian, Yang, and Cai]{tian_detecting_2018}
Shixin Tian, Guolei Yang, and Ying Cai.
\newblock Detecting {Adversarial} {Examples} {Through} {Image}
  {Transformation}.
\newblock \emph{Proceedings of the AAAI Conference on Artificial Intelligence},
  32\penalty0 (1), April 2018.
\newblock ISSN 2374-3468, 2159-5399.
\newblock \doi{10.1609/aaai.v32i1.11828}.

\bibitem[Touvron et~al.(2023)Touvron, Martin, Stone, Albert, Almahairi, Babaei,
  Bashlykov, Batra, Bhargava, Bhosale, et~al.]{touvron2023llama}
Hugo Touvron, Louis Martin, Kevin Stone, Peter Albert, Amjad Almahairi, Yasmine
  Babaei, Nikolay Bashlykov, Soumya Batra, Prajjwal Bhargava, Shruti Bhosale,
  et~al.
\newblock Llama 2: Open foundation and fine-tuned chat models.
\newblock \emph{arXiv preprint arXiv:2307.09288}, 2023.

\bibitem[Tramer(2022)]{tramer2022detecting}
Florian Tramer.
\newblock Detecting adversarial examples is (nearly) as hard as classifying
  them.
\newblock In \emph{International Conference on Machine Learning}, pp.\
  21692--21702. PMLR, 2022.

\bibitem[Tram{\`e}r et~al.(2018)Tram{\`e}r, Kurakin, Papernot, Goodfellow,
  Boneh, and McDaniel]{tramer2018ensemble}
Florian Tram{\`e}r, Alexey Kurakin, Nicolas Papernot, Ian Goodfellow, Dan
  Boneh, and Patrick McDaniel.
\newblock Ensemble adversarial training: Attacks and defenses.
\newblock In \emph{International Conference on Learning Representations}, 2018.

\bibitem[Tsipras et~al.(2018)Tsipras, Santurkar, Engstrom, Turner, and
  Madry]{tsipras2018robustness}
Dimitris Tsipras, Shibani Santurkar, Logan Engstrom, Alexander Turner, and
  Aleksander Madry.
\newblock Robustness may be at odds with accuracy.
\newblock In \emph{International Conference on Learning Representations}, 2018.

\bibitem[Tsuzuku et~al.(2018)Tsuzuku, Sato, and Sugiyama]{tsuzuku2018lipschitz}
Yusuke Tsuzuku, Issei Sato, and Masashi Sugiyama.
\newblock Lipschitz-margin training: Scalable certification of perturbation
  invariance for deep neural networks.
\newblock \emph{Advances in neural information processing systems}, 31, 2018.

\bibitem[Tsuzuku et~al.(2020)Tsuzuku, Sato, and
  Sugiyama]{tsuzuku2020normalized}
Yusuke Tsuzuku, Issei Sato, and Masashi Sugiyama.
\newblock Normalized flat minima: Exploring scale invariant definition of flat
  minima for neural networks using pac-bayesian analysis.
\newblock In \emph{International Conference on Machine Learning}, pp.\
  9636--9647. PMLR, 2020.

\bibitem[Virmaux \& Scaman(2018)Virmaux and Scaman]{virmaux2018lipschitz}
Aladin Virmaux and Kevin Scaman.
\newblock Lipschitz regularity of deep neural networks: analysis and efficient
  estimation.
\newblock \emph{Advances in Neural Information Processing Systems}, 31, 2018.

\bibitem[Walter et~al.(2022)Walter, Stutz, and Schiele]{walter_fragile_2022}
Nils~Philipp Walter, David Stutz, and Bernt Schiele.
\newblock On {Fragile} {Features} and {Batch} {Normalization} in {Adversarial}
  {Training}, April 2022.
\newblock arXiv:2204.12393 [cs, stat].

\bibitem[Wei et~al.(2024)Wei, Haghtalab, and Steinhardt]{wei2024jailbroken}
Alexander Wei, Nika Haghtalab, and Jacob Steinhardt.
\newblock Jailbroken: How does llm safety training fail?
\newblock \emph{Advances in Neural Information Processing Systems}, 36, 2024.

\bibitem[Wu et~al.(2020)Wu, Xia, and Wang]{wu2020adversarial}
Dongxian Wu, Shu-Tao Xia, and Yisen Wang.
\newblock Adversarial weight perturbation helps robust generalization.
\newblock \emph{Advances in neural information processing systems},
  33:\penalty0 2958--2969, 2020.

\bibitem[Xu et~al.(2020)Xu, Li, Jiang, and Xia]{xu2020adversarial}
Jia Xu, Yiming Li, Yong Jiang, and Shu-Tao Xia.
\newblock Adversarial defense via local flatness regularization.
\newblock In \emph{2020 IEEE International Conference on Image Processing
  (ICIP)}, pp.\  2196--2200. IEEE, 2020.

\bibitem[Xu et~al.(2018)Xu, Evans, and Qi]{xu_feature_2018}
Weilin Xu, David Evans, and Yanjun Qi.
\newblock Feature {Squeezing}: {Detecting} {Adversarial} {Examples} in {Deep}
  {Neural} {Networks}.
\newblock In \emph{Proceedings 2018 {Network} and {Distributed} {System}
  {Security} {Symposium}}, 2018.
\newblock \doi{10.14722/ndss.2018.23198}.
\newblock arXiv:1704.01155 [cs].

\bibitem[Yang et~al.(2020)Yang, Rashtchian, Zhang, Salakhutdinov, and
  Chaudhuri]{yang2020closer}
Yao-Yuan Yang, Cyrus Rashtchian, Hongyang Zhang, Russ~R Salakhutdinov, and
  Kamalika Chaudhuri.
\newblock A closer look at accuracy vs. robustness.
\newblock \emph{Advances in neural information processing systems},
  33:\penalty0 8588--8601, 2020.

\bibitem[Zagoruyko \& Komodakis(2016)Zagoruyko and
  Komodakis]{Zagoruyko2016WideRN}
Sergey Zagoruyko and Nikos Komodakis.
\newblock Wide residual networks.
\newblock \emph{British Machine Vision Conference}, abs/1605.07146, 2016.

\bibitem[Zou et~al.(2023)Zou, Wang, Kolter, and Fredrikson]{zou2023universal}
Andy Zou, Zifan Wang, J~Zico Kolter, and Matt Fredrikson.
\newblock Universal and transferable adversarial attacks on aligned language
  models.
\newblock \emph{arXiv preprint arXiv:2307.15043}, 2023.

\end{thebibliography}
